\def\year{2019}\relax
\documentclass[letterpaper]{article} %DO NOT CHANGE THIS
\usepackage{aaai19}  %Required
\usepackage{times}  %Required
\usepackage{helvet}  %Required
\usepackage{courier}  %Required
\usepackage{url}  %Required
\usepackage{graphicx}  %Required
\usepackage{subcaption}
\usepackage{multirow}
\usepackage{color}

\begin{document}
\title{Hotels-50K: A Global Hotel Recognition Dataset}
\author{Abby Stylianou\textsuperscript{1}, Hong Xuan\textsuperscript{1}, Maya Shende\textsuperscript{1},\\
\bf \Large Jonathan Brandt\textsuperscript{2}, Richard Souvenir\textsuperscript{3} and Robert Pless\textsuperscript{1} \\
\textsuperscript{1}George Washington University \\
\textsuperscript{2}Adobe Research\\
\textsuperscript{3}Temple University\\
{astylianou,xuanhong,mshende}@gwu.edu, jbrandt@adobe.com, souvenir@temple.edu, pless@gwu.edu
}
\maketitle
\begin{abstract}

Recognizing a hotel from an image of a hotel room is important for human trafficking investigations.  Images directly link victims to places and can help verify where victims have been trafficked, and where their traffickers might move them or others in the future.  Recognizing the hotel from images is challenging because of low image quality, uncommon camera perspectives, large occlusions (often the victim), and the similarity of objects (e.g., furniture, art, bedding) across different hotel rooms. 
To support efforts towards this hotel recognition task, we have curated a dataset of over 1 million annotated hotel room images from 50,000 hotels. These images include professionally captured photographs from travel websites and crowd-sourced images from a mobile application, which are more similar to the types of images analyzed in real-world investigations.  We present a baseline approach based on a standard network architecture and a collection of data-augmentation approaches tuned to this problem domain.  
\end{abstract}

\section{Introduction}
\begin{figure}
    \centering
    \includegraphics[width=.9\columnwidth]{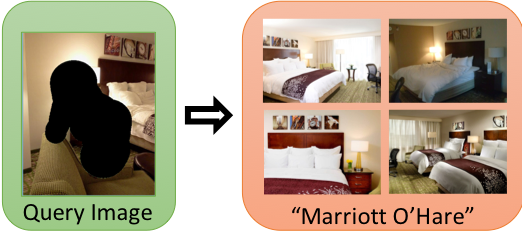}
    \caption{The Hotels-50K dataset supports the development of hotel recognition algorithms to help in investigations of human trafficking by identifying the hotel where a picture was taken.}
    \label{fig:frontPage}
\end{figure}

In recent years, the number of images of victims of human trafficking available online has grown at an alarming rate~\cite{bouche2015report,ncmecAmicusBrief}. Whether used
for advertising or exchanged among criminal networks, these photographs can serve as visual evidence of where the victim was trafficked. Such images are often captured in hotel rooms. Identifying the hotels in these photographs to understand where a victim was (Figure~\ref{fig:frontPage}), gives insight into trafficking operations, which is a a top priority for law enforcement~\cite{nationalStrategy}.

\begin{figure}
    \centering
    \includegraphics[height=1.8in]{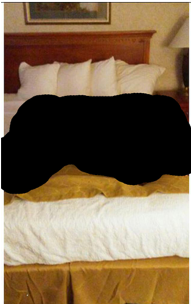}
    \includegraphics[height=1.8in]{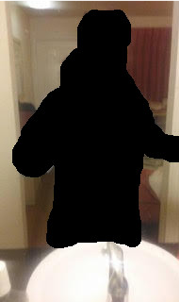}
    \includegraphics[height=1.8in]{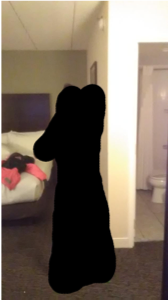}
    \caption{Example images from hotel rooms used in human trafficking investigations with the region containing the victim masked off.}
    \label{fig:queries}
\end{figure}

Figure~\ref{fig:queries} shows a few example of law enforcement queries.  Often the region of the images containing the victim is
masked for privacy and legal reasons.  Algorithms for recognition in this context must be robust to large occlusions, varying lighting conditions, and the unique perspectives of a hotel room.

This paper introduces the Hotels-50K dataset, which includes over 1 million images from 50,000 hotels around the world, designed to support efforts that address this challenging recognition task. Hotels-50K includes both professional photographs from travel websites and crowd-sourced images from a mobile application, which are more similar to the types of images analyzed in real-world investigations.  

This domain poses unique challenges compared to generic scene and place recognition tasks.  These recognition problems can be grouped based on the specificity of the categories~\cite{grauman_leibe_2011}:
\begin{enumerate}
    \item Basic-level categories (e.g., `building')
    \item Specialized categories (e.g., `church')
    \item Exact instances (e.g., `the Notre-Dame')
\end{enumerate}
The second task ("What type of building is this?") is often referred to as \emph{scene recognition} and the third task ("What specific church is this?") as \emph{place recognition}. Scene recognition requires learning the shared properties of the examples in the specialized class, while place recognition requires learning the specific components and their configuration that correspond to a particular instance.
Hotel recognition does not fit neatly into either task. It requires learning both the general, shared properties of all of the rooms in a particular hotel, such as its decor or star rating, or commonly used color profiles, as well as recognizing duplicated instances of furniture, art and bedding that may be used in different configurations throughout the hotel. 

This paper has three main contributions.  First, we propose and formulate the problem of hotel instance recognition.  Second, we curate and share a data set and evaluation protocol for this problem at a scale that is relevant to international efforts to address trafficking.  Third, we describe and test algorithms that include the data augmentation steps necessary to attack this problem as a reasonable baseline for comparisons.

\section{Related Work}
Hotels-50k is a large-scale dataset designed to support research in hotel recognition
for images with the long term goal of supporting robust applications to aid in criminal
investigations. In this section, we review related efforts towards (1) AI to combat human trafficking, (2) targeted large-scale image datasets, and (3) scene and place recognition.

\paragraph{AI to Combat Human Trafficking.} The Hotels-50K dataset and the problem of automatically recognizing hotel rooms fits within a larger set of efforts to apply machine learning, computer vision, and natural language processing to the domain of addressing human trafficking.  These efforts largely focused on indexing online escort advertisements, based on locations and phone numbers in the advertisement text or imprinted on advertising images~\cite{alvari2017semi,dubrawski2015leveraging,kejriwal2017investigative,szekely2015building}.  Additionally, there are larger-scale projects, such as Thorn\footnote{\url{https://www.wearethorn.org/}} that implement approaches including facial identification for identifying victims of child sex trafficking and sexual abuse.

\paragraph{Targeted Large-Scale Image Datasets} The computer vision community has a long tradition of developing datasets to support and challenge the research community. Some of most well-known datasets include ImageNet~\cite{deng2009imagenet}, Places~\cite{zhou2017places}, and CIFAR-100~\cite{krizhevsky2009learning}.
These benchmarks drive competitions for comparing classification and retrieval methods, but because they tend to focus on general (unrelated) categories of images there have been additional efforts towards curating domain-specific datasets, including datasets of classes of cars~\cite{CAR196} and birds~\cite{wah2011caltech}.  Most closely related to Hotels-50K are datasets that directly address investigative use-cases, including a database of tattoos~\cite{ngan2015tattoo}, and a dataset of advertisements labelled by whether they include a victim of trafficking~\cite{tong2017combating}. 

%moved this figure far forward to fix problems later.
\begin{figure}
    \centering
    \includegraphics[width=.95\columnwidth]{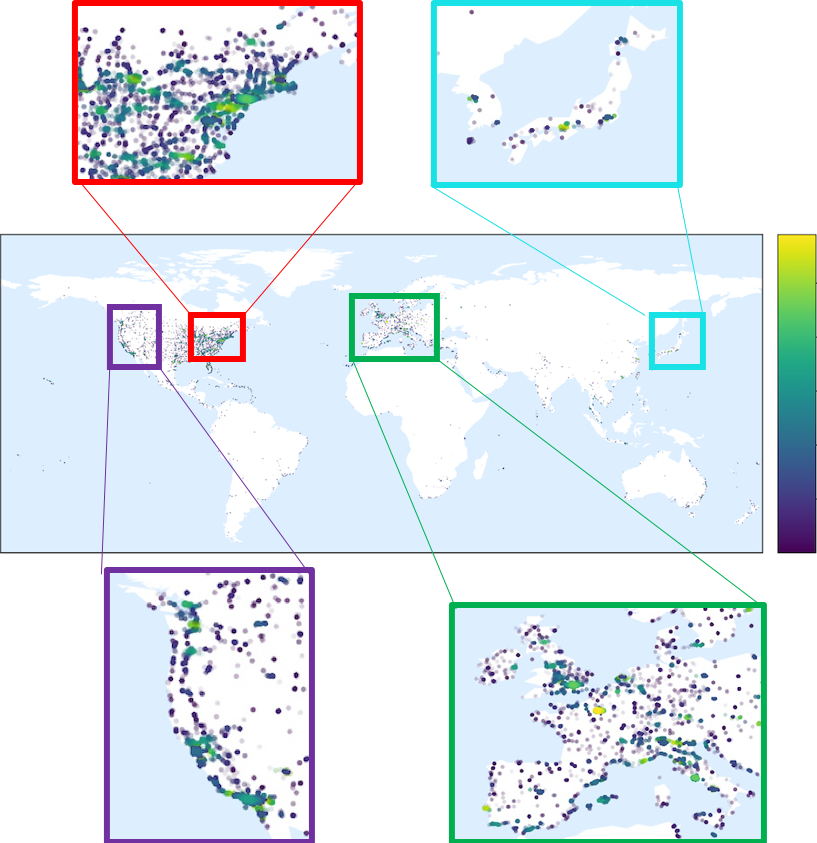}
    \caption{Geographic distribution of the Hotels-50K dataset, with a dot at every hotel location, color coded (from blue to yellow) by the local density of hotels. Images are most abundant in the United States, Western Europe and along popular coastlines. }
    \label{fig:geographicDensity}
\end{figure}

\paragraph{Scene and Place Recognition}
Recognizing the scene from which an image was captured has been a problem of great interest in the computer vision community. Most work in this area focuses on the problem of identifying the scene category (e.g., park, beach, parking lot) rather than particular locations, but recently there has been increased interest in estimating the precise geographic location of an image.

This place recognition problem can also be formulated as an image retrieval task where geotagged images serve as a database, and a query image's location is inferred by finding visually similar images in the dataset~\cite{baatz2012large,chen2011city,crandall2009mapping,hays2008im2gps,jacobs07geolocate,schindler2007city,torii2013visual,zamir2010accurate,googleLandmarks}. Increasingly, methods train deep neural networks to produce similar features for images from nearby locations~\cite{zhou2014recognizing,netvlad,visualPlaceRecognition,vo2017revisit,zhai2018geotemporal}.  

\newcommand{\exampleImWidth}{1in}
\newcommand{\exampleImHeight}{.65in}
\begin{figure*}
    \begin{tabular}{ccc|ccc}

        \multicolumn{3}{c|}{(a) Travel Websites} & \multicolumn{3}{c}{(b) TraffickCam} \\

        \raisebox{-.5\height}{\includegraphics[width=\exampleImWidth]{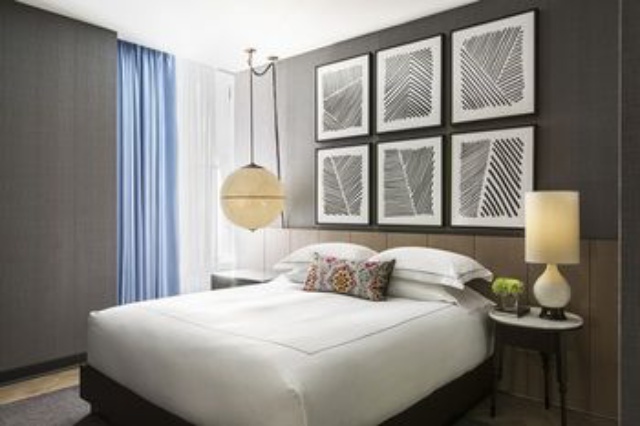}}
        &
        \raisebox{-.5\height}{\includegraphics[width=\exampleImWidth]{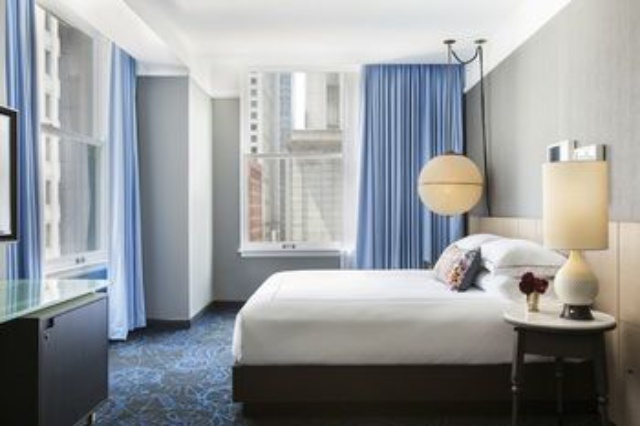}}
        &
        \raisebox{-.5\height}{\includegraphics[width=\exampleImWidth]{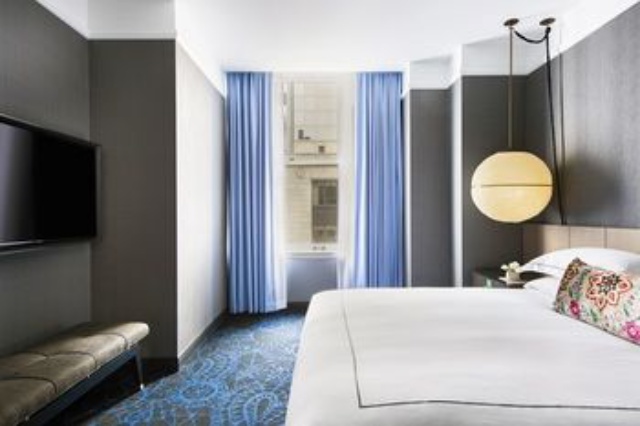}}
        &
        \raisebox{-.5\height}{\includegraphics[width=\exampleImWidth]{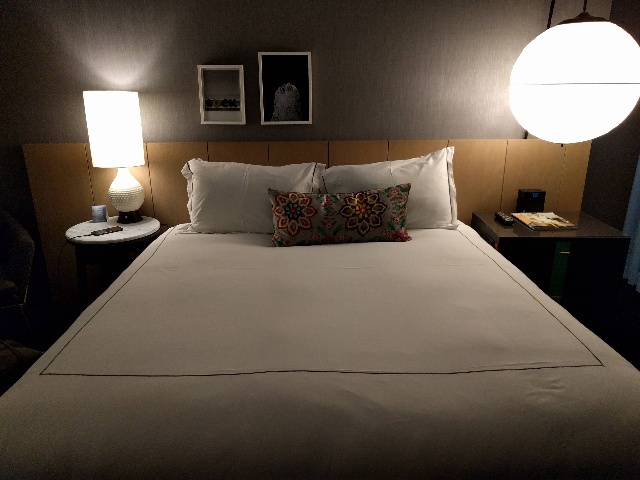}}
        & 
        \raisebox{-.5\height}{\includegraphics[width=.6in]{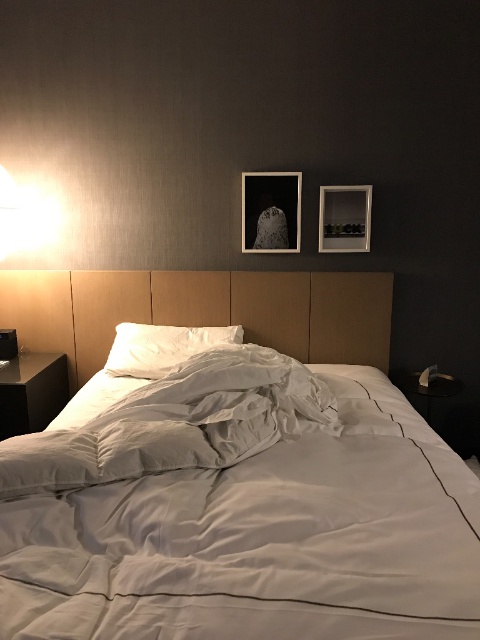}}
        & 
        \raisebox{-.5\height}{\includegraphics[width=.6in]{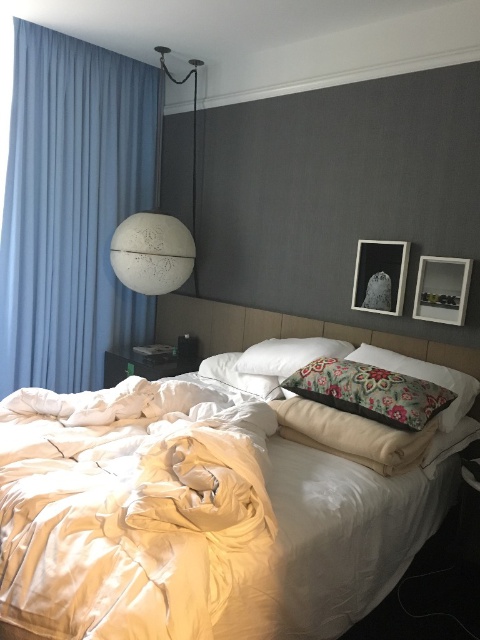}}
        
        \\
        \hline
        
        \raisebox{-.5\height}{\includegraphics[width=\exampleImWidth]{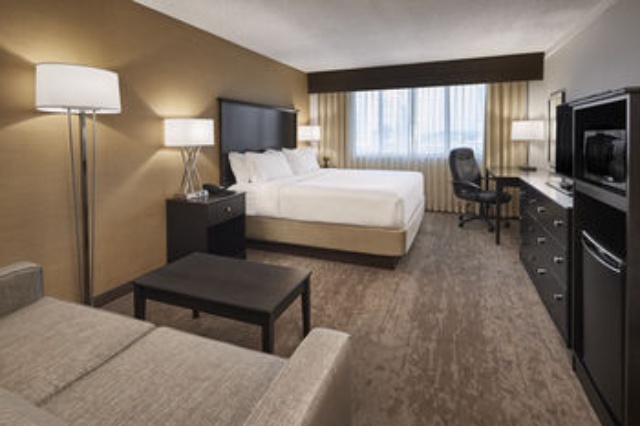}}
        &
        \raisebox{-.5\height}{\includegraphics[width=\exampleImWidth]{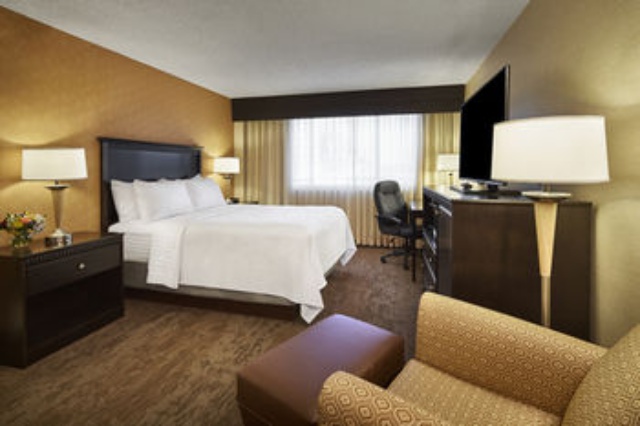}}
        &
        \raisebox{-.5\height}{\includegraphics[width=\exampleImWidth]{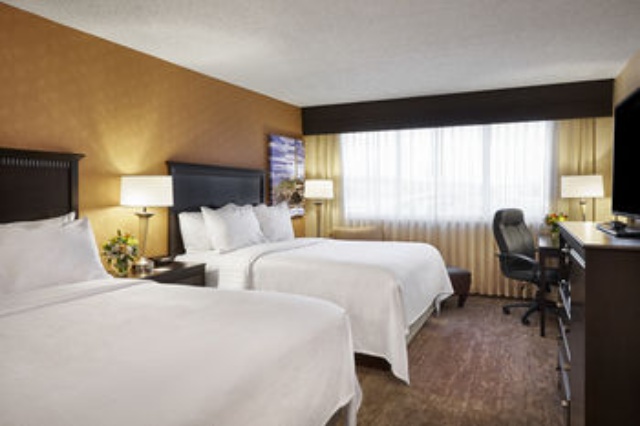}}
        &
        \raisebox{-.5\height}{\includegraphics[width=\exampleImWidth]{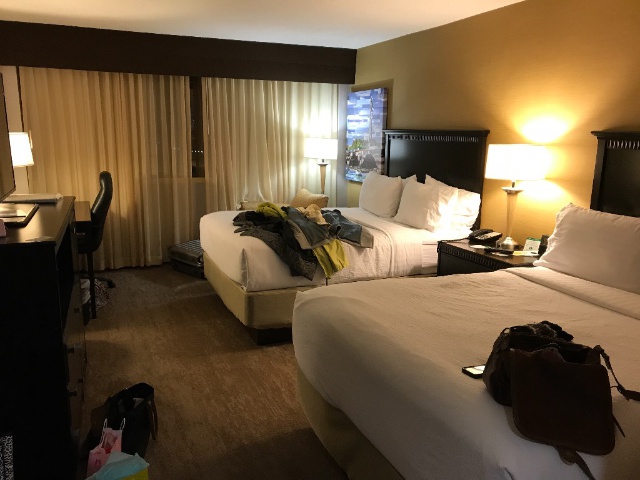}}
        & 
        \raisebox{-.5\height}{\includegraphics[width=.6in]{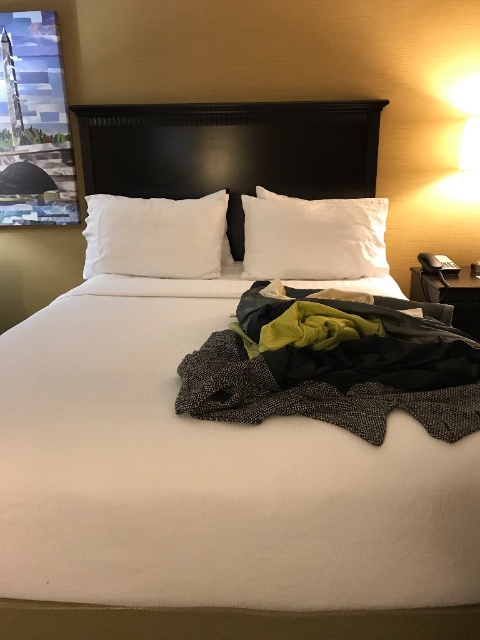}}
        & 
        \raisebox{-.5\height}{\includegraphics[width=.6in]{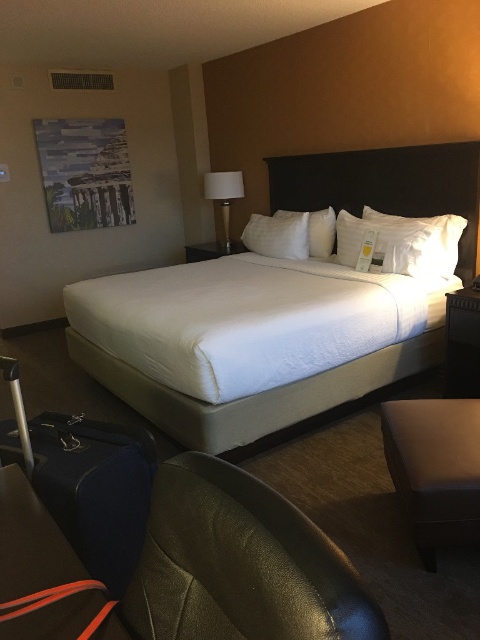}} 
        
        % \\
        % \hline

        % \raisebox{-.5\height}{\includegraphics[width=\exampleImWidth]{figures/example_images/expedia/1/1.png}}
        % & 
        % \raisebox{-.5\height}{\includegraphics[width=\exampleImWidth]{figures/example_images/expedia/1/3.png}}
        % & 
        % \raisebox{-.5\height}{\includegraphics[width=.65in]{figures/example_images/expedia/1/2.png}} 
        % &
        % \raisebox{-.5\height}{\includegraphics[width=\exampleImWidth]{figures/example_images/tcam/1/1.png}}
        % &
        % \raisebox{-.5\height}{\includegraphics[width=\exampleImWidth]{figures/example_images/tcam/1/2.png}}
        % &
        % \raisebox{-.5\height}{\includegraphics[width=.6in]{figures/example_images/tcam/1/3.png}} 
        
        \\
        \hline
        
        \raisebox{-.5\height}{\includegraphics[width=.96in]{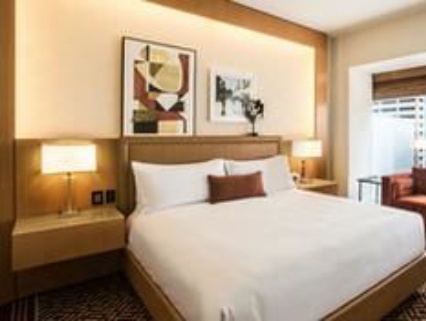}}
        &
        \raisebox{-.5\height}{\includegraphics[width=.95in]{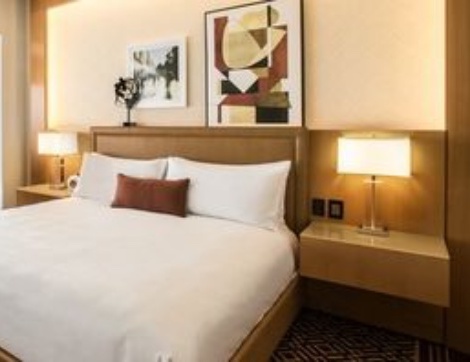}}
        &
        \raisebox{-.5\height}{\includegraphics[width=\exampleImWidth]{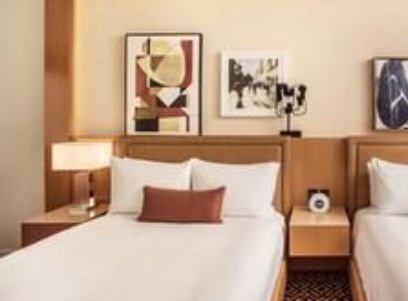}}
        &
        \raisebox{-.5\height}{\includegraphics[width=\exampleImWidth]{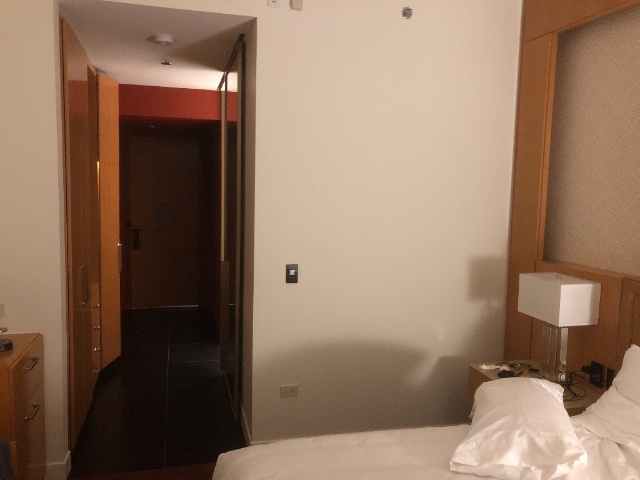}}
        &
        \raisebox{-.5\height}{\includegraphics[width=\exampleImWidth]{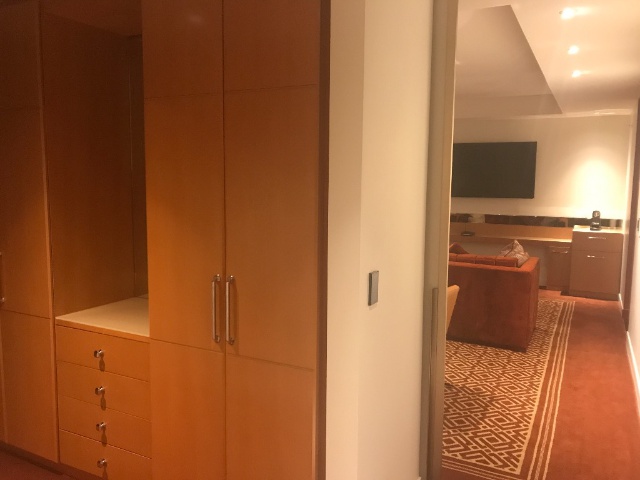}}
        &
        \raisebox{-.5\height}{\includegraphics[width=.6in]{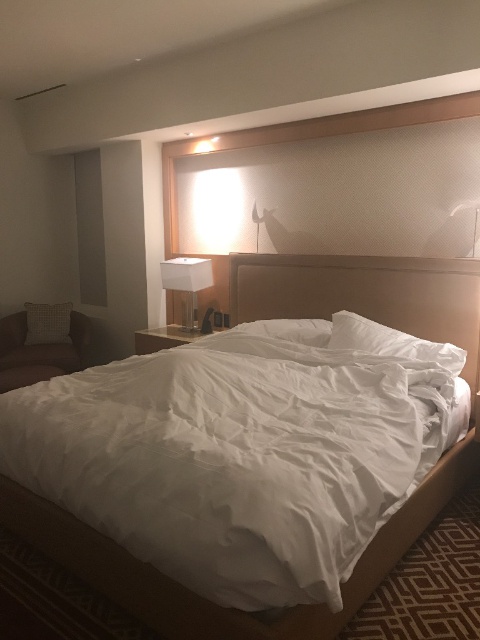}}
        
        \\
        \hline

        \raisebox{-.5\height}{\includegraphics[width=\exampleImWidth]{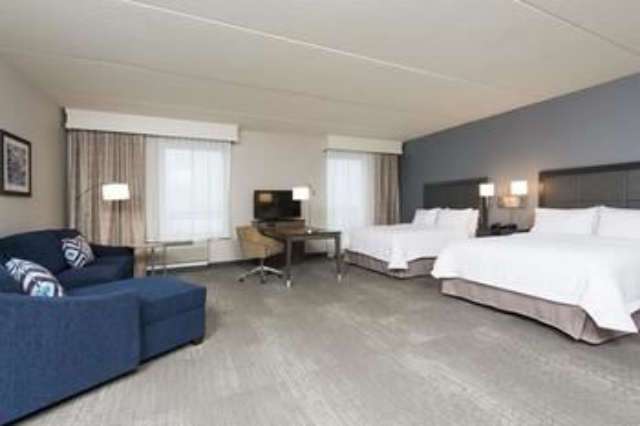}}
        &
        \raisebox{-.5\height}{\includegraphics[width=\exampleImWidth]{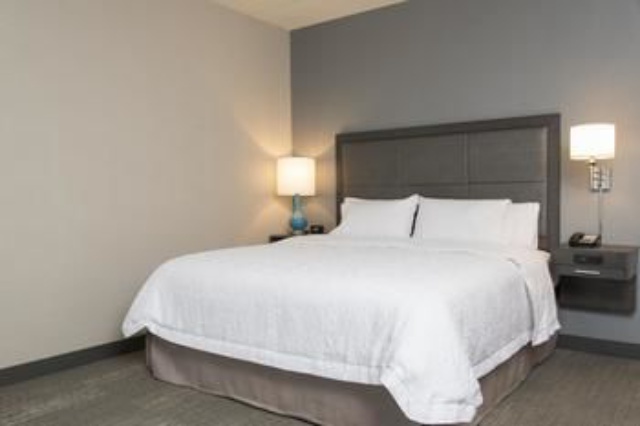}}
        &
        \raisebox{-.5\height}{\includegraphics[width=\exampleImWidth]{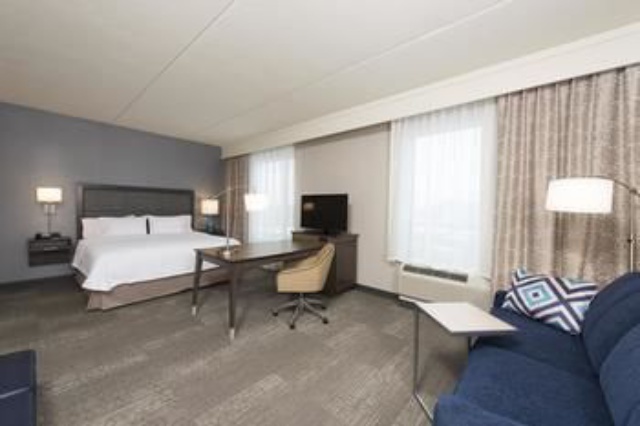}}
        & 
        \raisebox{-.5\height}{\includegraphics[width=\exampleImWidth]{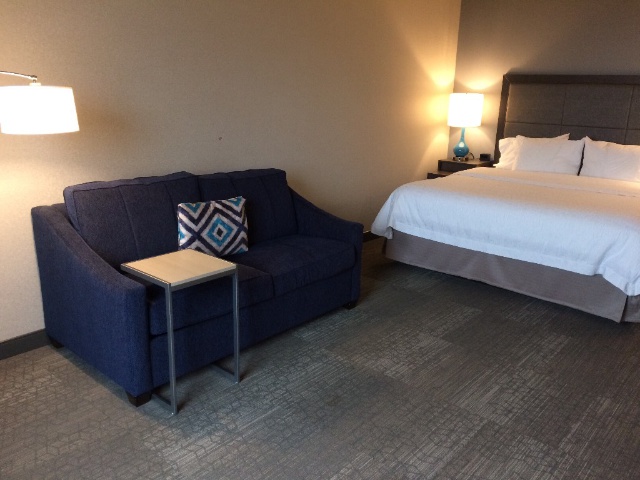}}
        &
        \raisebox{-.5\height}{\includegraphics[width=\exampleImWidth]{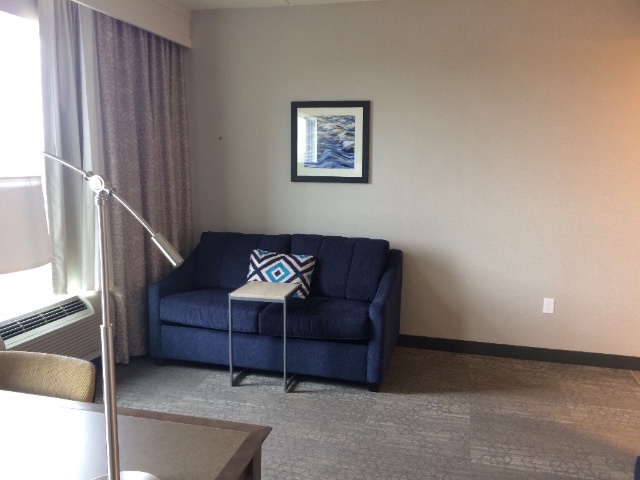}}
        & 
        \raisebox{-.5\height}{\includegraphics[width=.6in]{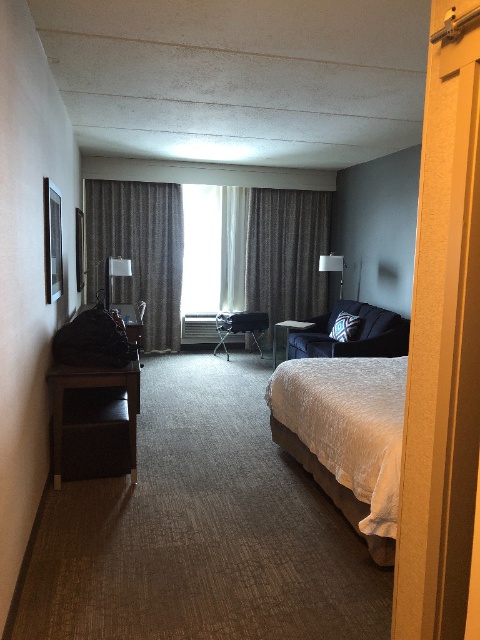}} 

    \end{tabular}
    \caption[Image variability from different sources]{Comparing images across data sources shows clear differences in image quality and lighting.  Each row shows images from the same hotel, with examples from (a) travel websites and (b) the TraffickCam crowd-sourcing app.}
    \label{fig:domainImages}
\end{figure*}

Algorithms trying recognize a specific place can exploit the fact that the same objects or landmarks appear in the same geometric configuration from different viewpoints. These geometric and matching approaches do not apply to hotel recognition.
Within a hotel, the rooms may have some objects that are the same (e.g., every room has the same headboard), some objects that are different (e.g., different artwork on the walls), and those objects may be in different configurations from room to room (e.g., two beds vs. one or furniture on different walls).

\paragraph{Summary} Hotels-50K follows in the tradition of large-scale datasets widely used in the computer vision and machine learning communities. This dataset will support and complement the recent trend for using AI to combat criminal activity, specifically human trafficking. The problem of hotel recognition poses unique challenges and existing methods designed for recognizing outdoor scenes or landmarks are not well-suited to the problem of discriminating between similar-looking hotel rooms. 

\section{The Hotels-50K Dataset}

Hotels-50K consists of 1,027,871 images from 50,000 unique hotels around the world. 
Each of the images in the Hotels-50K dataset includes the following metadata: (1) hotel name
(2) geographic location, and (3) hotel chain, or \emph{Other} if the hotel property is
not part of a major chain. Figure~\ref{fig:geographicDensity} shows the geographic distribution of the images in our dataset. While the dataset consists of images from around the world, the images are more densely captured in the United States, Western Europe, and coastal regions.

\paragraph{Data Sources} The images in Hotels-50K come from two primary sources: (1) scraped from publicly 
available travel websites, such as Expedia and (2) captured by the crowdsourcing 
mobile application, TraffickCam, which allows travelers to submit photos of their 
hotel room. Figure~\ref{fig:domainImages} shows example images from both sources captured at the same hotel. The photos from the travel websites 
are abundant, accounting for a majority of the images in the dataset. However, these images tend to be taken for 
promotional purposes, by professional photographers with excellent lighting 
conditions, of the nicest rooms in a hotel. These images are visually quite different from the types of images referenced in human 
trafficking investigations. On the other hand, while there are fewer crowdsourced images, these share more visual characteristics with the images used in real-world queries. The crowdsourced images are taken similar devices, at varying orientations, with luggage and other clutter, and without professional lighting.

\paragraph{Dataset Statistics} Of the 50,000 hotel classes in the Hotels-50K training dataset, 13,900 have TraffickCam user-submitted images (a total of 55,061 TraffickCam images are included in the training set). There are no hotels in the dataset that have only TraffickCam images.

\begin{figure*}
    \centering    \includegraphics[width=2\columnwidth]{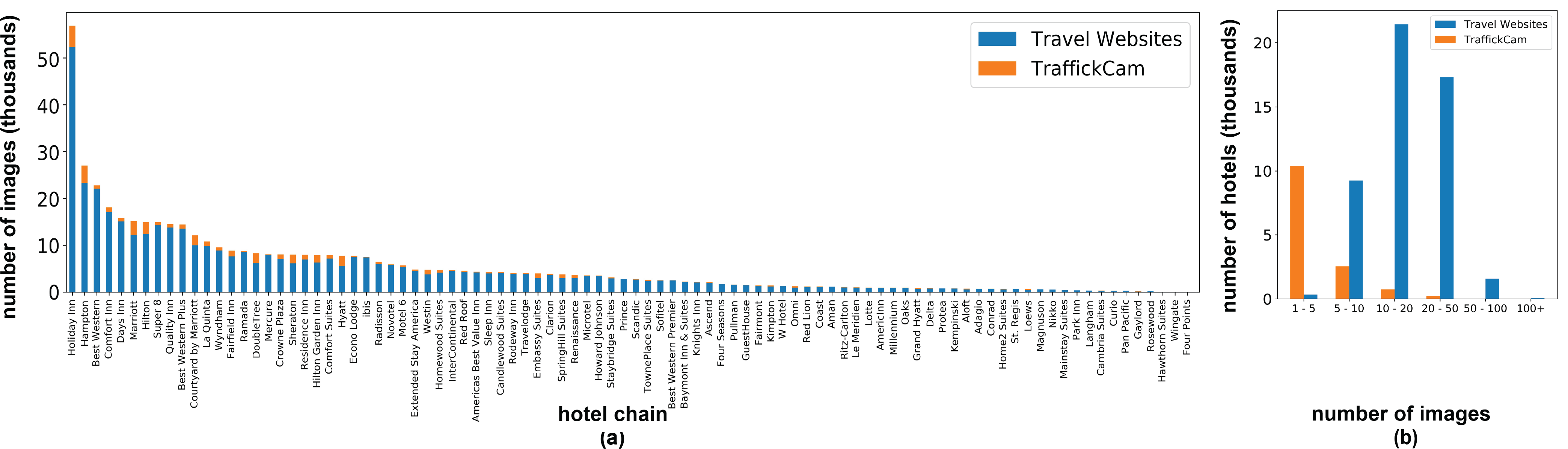}
    \caption{(a) Number of images, by source, for each of the 92 chains represented in the Hotels-50K dataset. (b) Histogram of the number of images per hotel in the Hotels-50K dataset, by the source.}
    \label{fig:imsPerClass}
\end{figure*}

Figure~\ref{fig:imsPerClass} show two histograms that characterize the sampling in the dataset.  Figure~\ref{fig:imsPerClass}(a) shows the number of images per hotel chain for each of the 92 major hotel chains represented in the Hotels-50K dataset.  Some chains have many more images than others (Holiday Inn, Hampton and Best Western), consistent with the prevalence of those hotel chains around the world.  Figure~\ref{fig:imsPerClass}(b) shows a histogram of the number of images per hotel broken down by the source of images (travel websites or TraffickCam mobile application). The average number of images from travel websites per hotel is $19.5$. The average number of images from TraffickCam for the hotels with TraffickCam images is $4.0$.

\newcommand{\chnConfusionHeight}{.65in}
\begin{figure}[t]
    \centering
    \begin{tabular}{lccc}
    \raisebox{-.5\height}{\rotatebox{90}{Motel 6}}&\raisebox{-.5\height}{\includegraphics[height=\chnConfusionHeight]{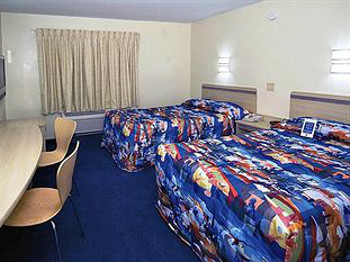}} & \raisebox{-.5\height}{\includegraphics[height=\chnConfusionHeight]{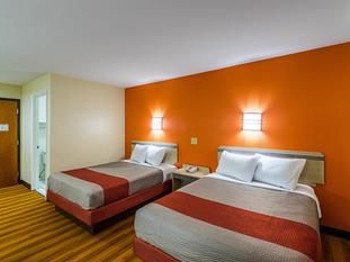}} & \raisebox{-.5\height}{\includegraphics[height=\chnConfusionHeight]{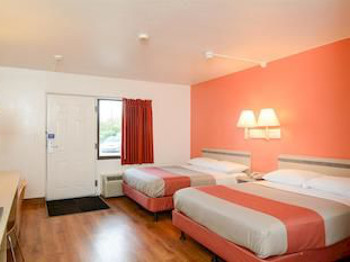}}
    \\
    \\
    \raisebox{-.5\height}{\rotatebox{90}{Super 8}}&\raisebox{-.5\height}{\includegraphics[height=\chnConfusionHeight]{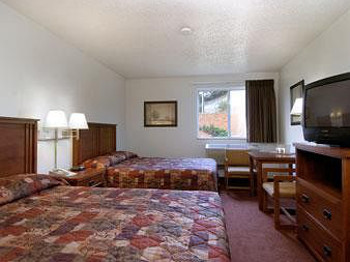}} & \raisebox{-.5\height}{\includegraphics[height=\chnConfusionHeight]{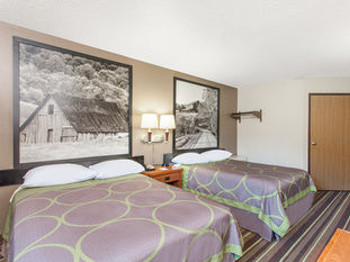}} & \raisebox{-.5\height}{\includegraphics[height=\chnConfusionHeight]{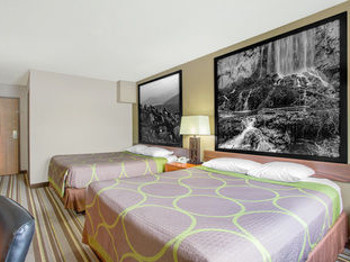}}
    \\
    \\
    \raisebox{-.5\height}{\rotatebox{90}{Extended Stay}}&\raisebox{-.5\height}{\includegraphics[height=\chnConfusionHeight]{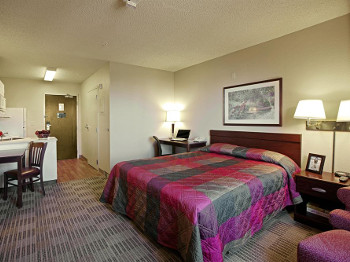}} & \raisebox{-.5\height}{\includegraphics[height=\chnConfusionHeight]{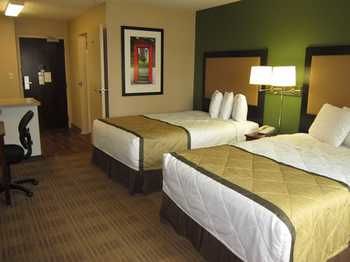}} & \raisebox{-.5\height}{\includegraphics[height=\chnConfusionHeight]{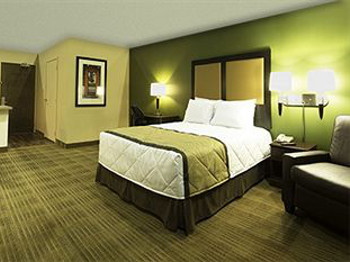}}
    \\
    \end{tabular}
    \caption{In each row, the first two images are from the same hotel, and the third is from a different hotel of the same chain.  This highlights one of the main challenges with hotel recognition, that images within the same hotel may be visually dissimilar, while images from different hotels, especially those from the same chain, may be visually similar.  }
    \label{fig:sameHotel_vs_sameChain}
\end{figure}

\paragraph{Observations} While there exist discriminative patterns and unique features visible
in the images from the hotels in Hotels-50K, this dataset highlights one of the main challenges in
hotel recognition. There can be high intraclass variation, as not every room within a single hotel will have the same shared properties or objects -- some rooms contain more amenities and some may have been renovated. On the other hand, there can be low interclass variation, especially from hotels of the same chain, making the recognition of a specific hotel difficult.  Figure~\ref{fig:sameHotel_vs_sameChain} shows a few specific examples where two rooms in the same hotel look much more different than rooms in two different hotels from the same chain.

\section{Evaluation Protocol}
Hotels-50K includes a separate test set of images to support the consistent evaluation of 
algorithms. Obtaining a large collection of images from real-world investigations
is problematic for many reasons. However, in the images in the test set are
meant to replicate, as closely as possible the type of data used in these cases.

The test set consists of 17,954 images from the TraffickCam mobile application from 5,000 different 
hotels, which are a subset of those found in the training set. There is no overlap in the 
mobile app users between the training and testing sets to avoid the case of near
duplicates due to multiple images from the same user with the same device captured at the same time.

\begin{figure}
    \centering
    % \begin{subfigure}[b]{.45\columnwidth}
    %     \centering
    %     \includegraphics[width=.95\columnwidth]{figures/example_masks/0.jpg}
    %  %   \caption{5-25\% occluded}
    % \end{subfigure}
    \begin{subfigure}[b]{.49\columnwidth}
        \centering
        \includegraphics[width=.95\columnwidth]{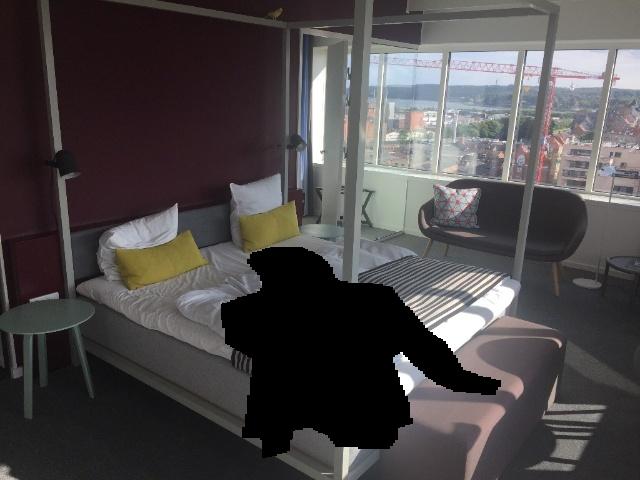}
      %  \caption{25-45\% occluded}
    \end{subfigure}
    % \begin{subfigure}[b]{.45\columnwidth}
    %     \centering
    %     \includegraphics[width=.95\columnwidth]{figures/example_masks/2.jpg}
    %   % \caption{45-65\% occluded}
    % \end{subfigure}
    \begin{subfigure}[b]{.49\columnwidth}
        \centering
        \includegraphics[width=.95\columnwidth]{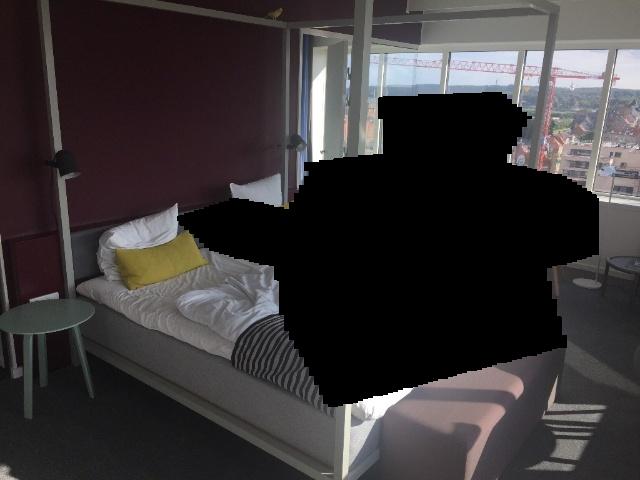}
        %\caption{65-85\% occluded}
    \end{subfigure}
    \caption{The images in the test set are augmented with person-shaped masks of varying size.}
    \label{fig:example_masks}
\end{figure}

To replicate real-world conditions where the regions of the image containing victims are 
masked prior to image analysis, the images are augmented with increasingly 
larger "people-shaped" masks. The masks are generated using silhouettes from 'people' regions 
in the MS-COCO semantic labels dataset~\cite{mscoco}. There are four levels of masking (none, low, medium high), corresponding to the relative sizes of the masked region in each image, where the largest masks can occupy up to 85\% of the height of the image. Figure~\ref{fig:example_masks} shows examples of masked test images.

The evaluation consists of the following tasks:
\begin{description}
\item[Hotel Instance Recognition] The goal for this task is to identify the hotel instance represented for each of the images in the test set.
\item[Hotel Chain Recognition]
The goal for this task is to identify the hotel chain represented in the image. Of 
the test set, 13,136 images are from one of 88 major hotel chains, with the remainder in 
the "Other" category.
\end{description}

\subsection{Evaluation Metrics}
Hotel recognition can be framed as both a classification task (i.e., predict the label given the image) and a retrieval task (i.e., find the most similar database images to a query). The evaluation suite for Hotels-50K supports both variants.

For the retrieval variant, the results should be provided as a
ranked list of the IDs of the 100 most similar images from the Hotels-50K dataset 
to each of the test images. The evaluation metric is top-$K$ accuracy, with $K = \{1, 10, 100\}$ for 
hotel instance recognition and $K = \{1, 3, 5\}$ for hotel chain recognition.

For the classification variant, the results should be provided 
as the posterior probabilities of hotel chains or instances
for each of the test images. The evaluation metrics include the
average multi-class log loss (lower is better) and top-$K$ 
classification accuracy with $K = \{1, 10, 100\}$ for hotel 
instance recognition and $K = \{1, 3, 5\}$ for hotel chain 
recognition.

\begin{figure}
\centering
   \begin{subfigure}[b]{.19\columnwidth}
        \centering
        \includegraphics[width=.98\columnwidth]{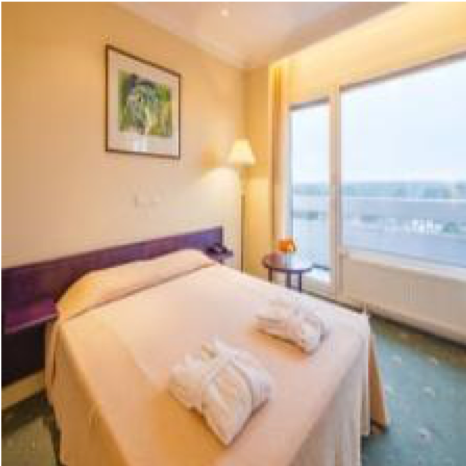}
        \caption{}
    \end{subfigure}
   \begin{subfigure}[b]{.19\columnwidth}
        \centering
        \includegraphics[width=.98\columnwidth]{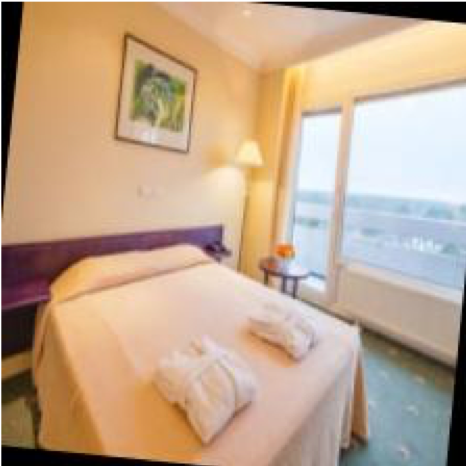}
        \caption{}
    \end{subfigure}
    \begin{subfigure}[b]{.19\columnwidth}
        \centering
        \includegraphics[width=.98\columnwidth]{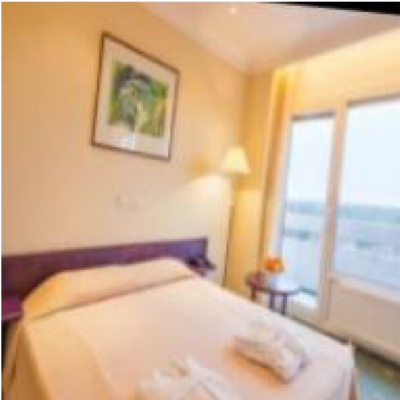}
        \caption{}
    \end{subfigure}
    \begin{subfigure}[b]{.19\columnwidth}
        \centering
        \includegraphics[width=.98\columnwidth]{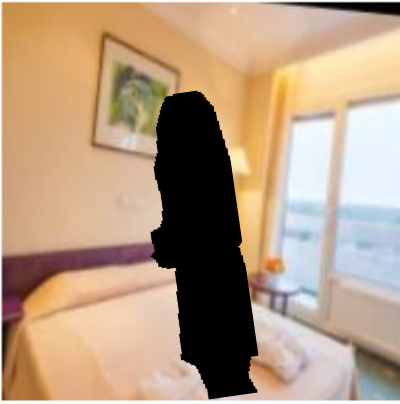}
        \caption{}
    \end{subfigure}
    \begin{subfigure}[b]{.19\columnwidth}
        \centering
        \includegraphics[width=.98\columnwidth]{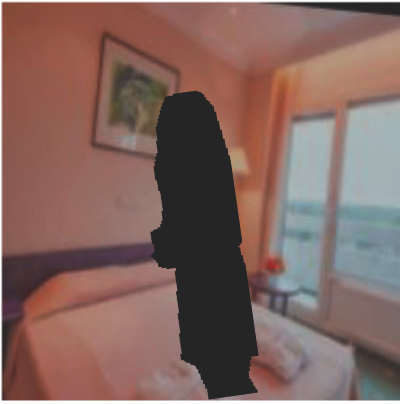}
        \caption{}
    \end{subfigure}
    \caption[Data augmentation to mimic properties of query images.]{Data augmentation steps to better match across different lighting conditions, scales and perspectives, and in the presence of large occlusions: (a) the original image; (b) after rotation; (c) after cropping; (d) after people mask applied; (e) after color filter rendered.}
    \label{fig:data_augmentation}
\end{figure}

\section{Results}
In order to set the baseline for performance on the Hotels-50K 
dataset, we compare two "off-the-shelf" pre-trained networks 
trained for object and scene
recognition to a method using data and augmentation schemes specifically tailored to hotel recognition. 

\subsection{Models}
For the pretrained models, we use the fixed feature representations and refer to these as the {\sc Fixed-Object} and {\sc Fixed-Scene} methods. The {\sc Fixed-Object} method is a Resnet-50 network trained on ImageNet (ILSVRC-2012)~\cite{resnet,deng2009imagenet,ILSVRC15}. The feature representation is the 1001-dimensional output from the final fully connected layer. The {\sc Fixed-Scene} method uses a VGG model trained on the Places365 dataset~\cite{zhou2017places}. The feature representation is the 512-dimensional output of the final pooling layer.

Our method uses the Hotels-50K training set as input
to fine tune a Resnet-50 model, pre-trained for ImageNet, to output 256-D features. The training scheme is the  combinatorial variant of triplet loss described in~\cite{HermansBeyer2017Arxiv}.

In training, we balance the number of crowdsourced and travel website images in each batch. Additionally, we perform a set of data augmentation steps, highlighted in Figure~\ref{fig:data_augmentation}. Images from the batch are randomly selected and rotated between -35 and 35 degrees, cropped between 60\% and 100\% of the original size, modified with color and brightness, and masked with person shaped silhouettes, similar to process used for the test data. The set of masks applied in training do not overlap with those used to generated the Hotels-50K test data and will be made available. Training parameters
were selected using cross-validation. The final model
was fine-tuned for 65,000 iterations with 120 images per batch.

\subsection{Retrieval}
For retrieval, we compute feature representations
for all of the images in the Hotels-50K training set using
each method. Feature representations are also computed 
for each image in the test set, and the database images
are ranked by cosine similarity to each test image.

 \begin{table}
    \centering
      \begin{tabular}{c|ccc|}
      \multicolumn{1}{c}{} & \multicolumn{3}{c}{\textbf{Instance}} \\
      \cline{2-4}
      & K=1 & 10 & 100 \\
      \cline{1-4}
      \multicolumn{1}{|c|}{{\sc Fixed-Object}} & 0.8 & 0.9 & 1.3 \\
      \multicolumn{1}{|c|}{{\sc Fixed-Scene}} & 0.2 & 0.8 & 2.4\\
      \multicolumn{1}{|c|}{Ours} & \textbf{8.1} & \textbf{17.6} & \textbf{34.8} \\
      \cline{1-4}
      \multicolumn{4}{c}{} \\
      \multicolumn{1}{c}{} & \multicolumn{3}{c}{\textbf{Chain}} \\
      \cline{2-4}
      & K=1 & 3 & 5 \\
      \cline{1-4}
      \multicolumn{1}{|c|}{{\sc Fixed-Object}} & 5.0 & 29.0 & \textbf{79.2} \\
      \multicolumn{1}{|c|}{{\sc Fixed-Scene}} & 7.2 & 34.2 & 78.7\\
      \multicolumn{1}{|c|}{Ours} & \textbf{42.5} & \textbf{56.4} & 62.8 \\
      \cline{1-4}
      
      \end{tabular}
      \caption{Retrieval results by hotel instance and by hotel chain, reported as top-$K$ accuracy.}
      \label{tab:hotel_vs_chain_comparison}
\end{table}

\begin{table*}
    \centering
    \def\arraystretch{1.1}
    \begin{tabular}{l|ccc|ccc|ccc|ccc|ccc|}
        \multicolumn{1}{r}{\textbf{Occlusion:}}&\multicolumn{3}{c}{\textbf{none}} & \multicolumn{3}{c}{\textbf{low}} & \multicolumn{3}{c}{\textbf{medium}} & \multicolumn{3}{c}{\textbf{high}} \\
        \cline{2-13}
    %     & & 
    %     K=1 & 3 & 5 &
    %     1 & 3 & 5 & 
    %     1 & 3 & 5 & 
    %     1 & 3 & 5 \\
    %     \cline{2-14}
    %     \multirow{3}{*}{\raisebox{-.5\height}{\rotatebox{90}{Chain}}} & \multicolumn{1}{|l|}{{\sc Fixed-Object}} &
    %     5.0 & 29.0 & 79.2 &
    %     5.0 & 29.0 & 79.2 &
    %     5.0 & 29.0 & 79.2 & 
    %     5.0 & 29.0 & 79.2\\
    %     & \multicolumn{1}{|l|}{{\sc Fixed-Scene}} & 
    %     7.2 & 34.2 & 78.7 &
    %     7.2 & 34.2 & 78.7 & 
    %     7.2 & 34.2 & 78.7 & 
    %     7.2 & 34.1 & 78.7 \\
    %     & \multicolumn{1}{|l|}{Ours} & 
    %     \textbf{39.7} & \textbf{68.6}  & \textbf{90.6} & 
    %     \textbf{39.7} & \textbf{68.6}  & \textbf{90.6}  & 
    %     \textbf{39.7} & \textbf{68.6}  & \textbf{90.6}  & 
    %     \textbf{39.7} & \textbf{68.6}  & \textbf{90.6}  \\
    %     \cline{2-14}
    %     & \multicolumn{13}{c}{}
    % \\
    %     \cline{3-14}
        & K=1 & 10 & 100 &
        1 & 10 & 100 & 
        1 & 10 & 100 & 
        1 & 10 & 100 \\
        \cline{1-13}
        \multicolumn{1}{|l|}{{\sc Fixed-Object}} & 
        0.8 & 0.9 & 1.3 &
        0.3 & 0.4 & 0.7 & 
        0.0 & 0.0 & 0.0 & 
        0.0 & 0.1 & 0.4 \\
        \multicolumn{1}{|l|}{{\sc Fixed-Scene}} & 
        0.2 & 0.8 & 2.4 &
        0.1 & 0.5 & 1.9 & 
        0.1 & 0.4 & 1.5 & 
        0.0 & 0.1 & 1.0 \\
        \multicolumn{1}{|l|}{Ours} & 
        \textbf{8.1} & \textbf{17.6} & \textbf{34.8} &
        \textbf{7.1} & \textbf{16.4} & \textbf{33.1} & 
        \textbf{5.9} & \textbf{14.1} & \textbf{29.9} & 
        \textbf{4.2} & \textbf{10.5} & \textbf{24.0} \\
        \cline{1-13}
    \end{tabular}
    \caption{Image retrieval comparison reported as top-$K$ accuracy.}
    \label{tab:topK_accuracy}
\end{table*}

Table~\ref{tab:hotel_vs_chain_comparison} shows the image retrieval results by hotel instance and chain for all three methods. For all methods, the retrieval accuracy by hotel instance is significantly lower than the accuracy by hotel chain.  This is likely due to the difficulty discriminating between particular instances of hotel chains that look similar. The chain identification task is simple enough that even the fixed methods not fine-tuned to the task achieve
nearly 80\% top-5 accuracy on this task. Therefore, for our remaining
experiments, we focus on the more challenging problem to recognize a hotel instance.

Table~\ref{tab:topK_accuracy} shows the image retrieval results for all three methods for the test images with varying sizes of image masking. Our approach has significantly higher retrieval accuracy compared to the pre-trained approaches for all tests, both with and without occlusions.

\newcommand{\exResultsHeight}{.56in}
\newcommand{\queryImWidth}{.15\textwidth}
\begin{figure*}
\setlength{\tabcolsep}{1pt}
    \begin{tabular}{cccccccccccc}
    Query Image & Model & 1 & 2 & 3 & 4 & 5\\
    \cline{1-7}
        \multirow{3}{*}{\raisebox{-.75\height}{ \includegraphics[width=\queryImWidth]{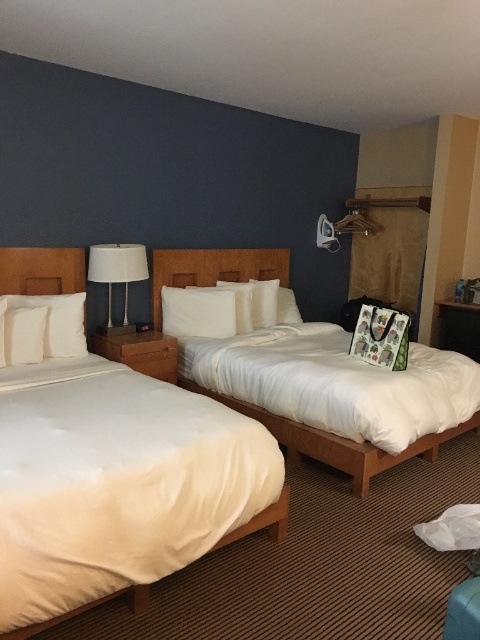}}}
        &
        \raisebox{-.5\height}{{\sc Fixed-Object}}
        & \raisebox{-.5\height}{\includegraphics[height=\exResultsHeight]{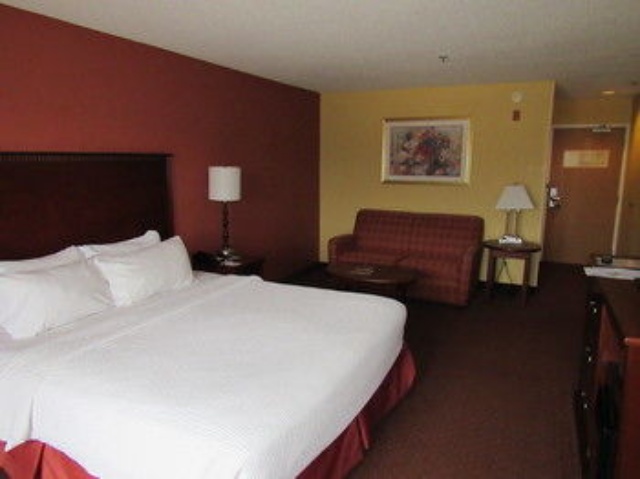}}
        &\raisebox{-.5\height}{\includegraphics[height=\exResultsHeight]{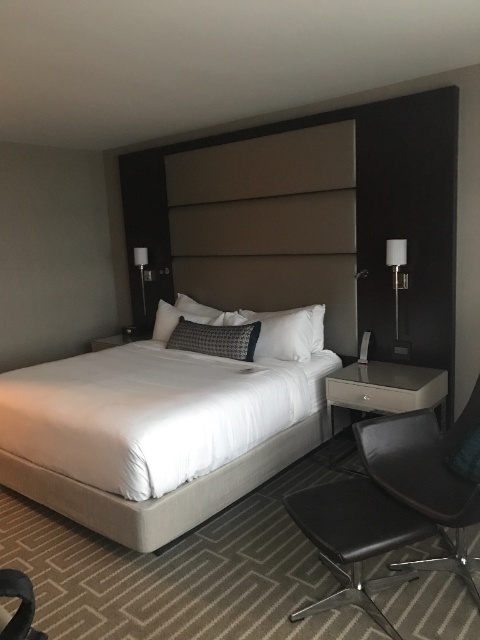}}
        &\raisebox{-.5\height}{\includegraphics[height=\exResultsHeight]{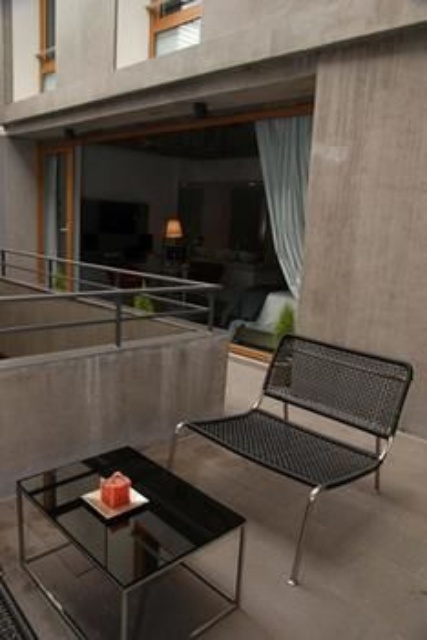}}
        &\raisebox{-.5\height}{\includegraphics[height=\exResultsHeight]{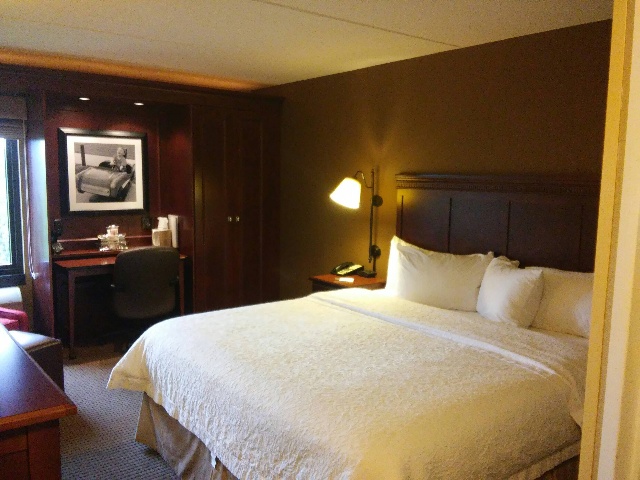}}
        &\raisebox{-.5\height}{\includegraphics[height=\exResultsHeight]{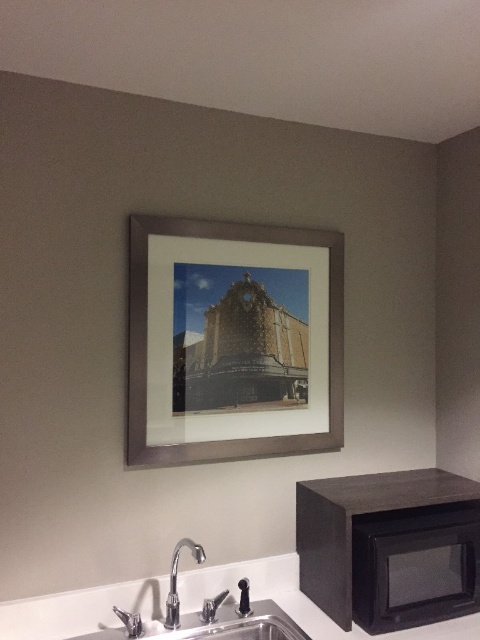}}
    \\
    &
        \raisebox{-.5\height}{{\sc Fixed-Scene}}
        &\raisebox{-.5\height}{\includegraphics[height=\exResultsHeight]{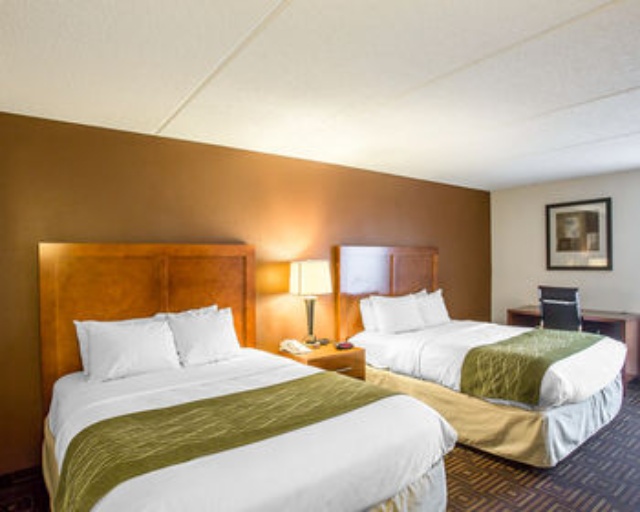}}
        &\raisebox{-.5\height}{\includegraphics[height=\exResultsHeight]{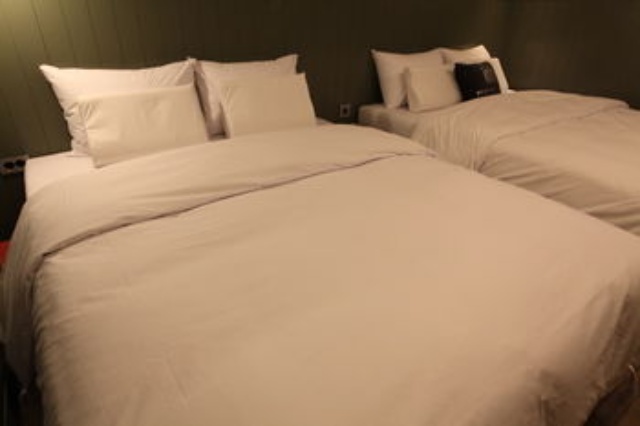}}
        &\raisebox{-.5\height}{\includegraphics[height=\exResultsHeight]{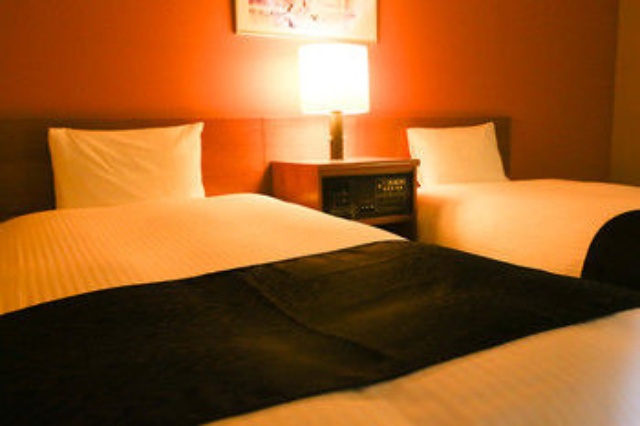}}
        &\raisebox{-.5\height}{\includegraphics[height=\exResultsHeight]{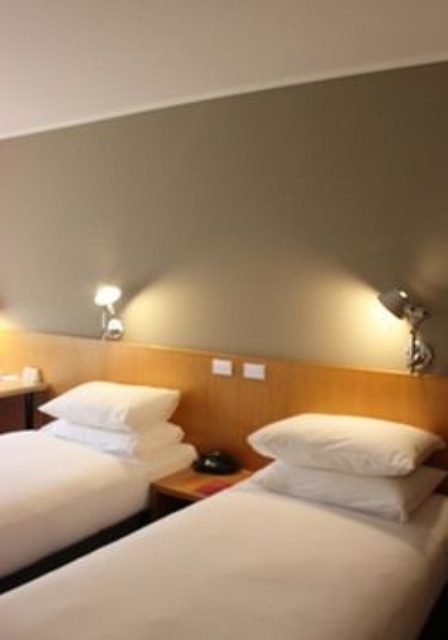}}
        &\raisebox{-.5\height}{\includegraphics[height=\exResultsHeight]{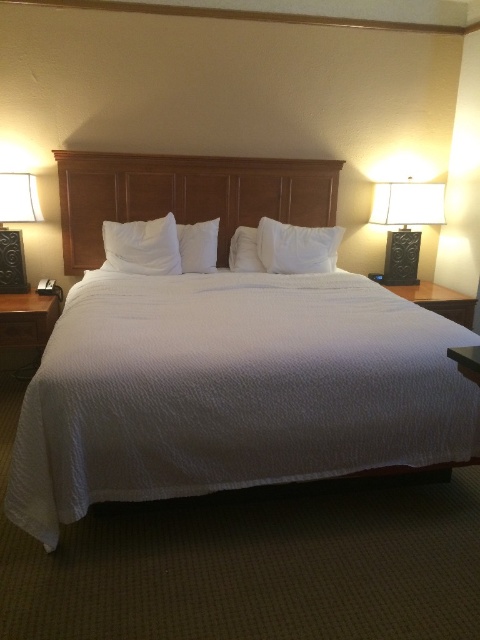}}
    \\
    &
        \raisebox{-.5\height}{Ours}
        & \raisebox{-.5\height}{\fcolorbox{green}{green}{\includegraphics[height=\exResultsHeight]{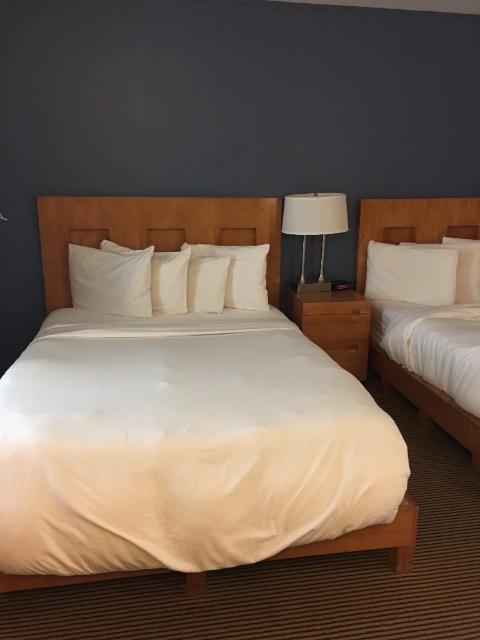}}}
        &\raisebox{-.5\height}{\fcolorbox{green}{green}{\includegraphics[height=\exResultsHeight]{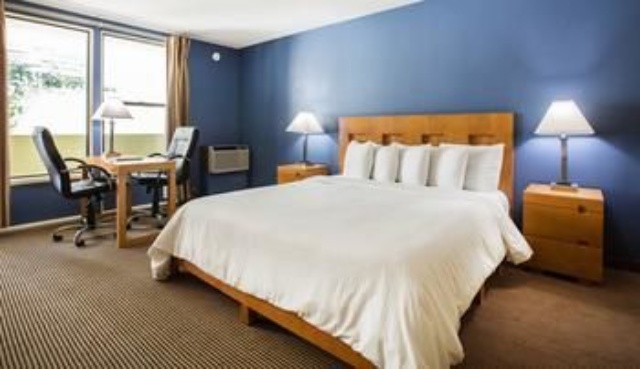}}}
        &\raisebox{-.5\height}{\includegraphics[height=\exResultsHeight]{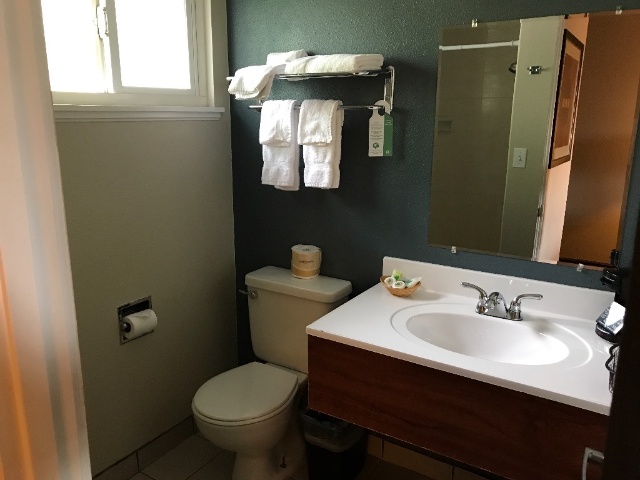}}
        &\raisebox{-.5\height}{\includegraphics[height=\exResultsHeight]{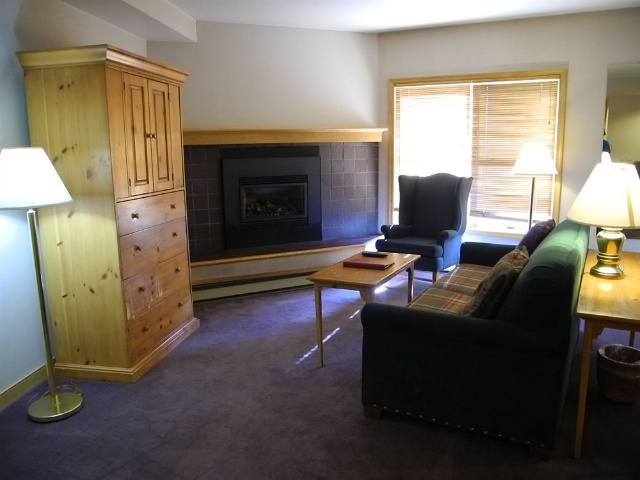}}
        & \raisebox{-.5\height}{\includegraphics[height=\exResultsHeight]{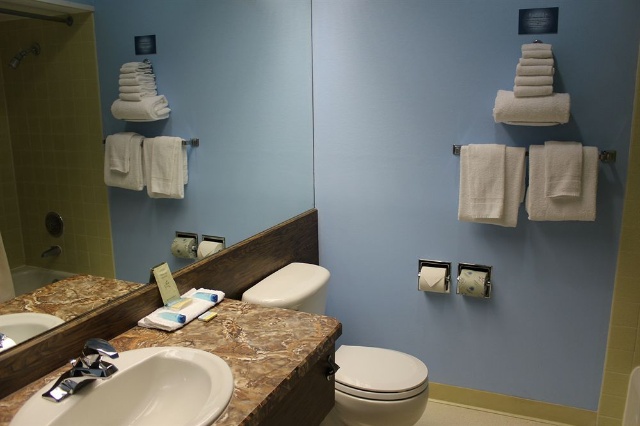}}\\
    \cline{1-7}
    
    \multirow{3}{*}{\raisebox{-1.2\height}{\includegraphics[width=\queryImWidth]{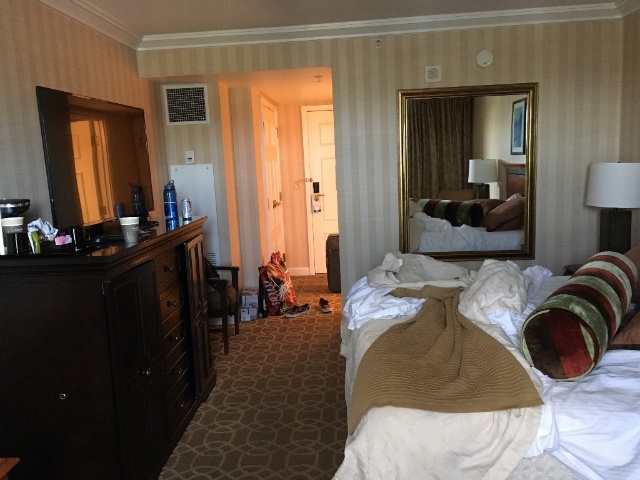}}}
        &\raisebox{-.5\height}{{\sc Fixed-Object}}
        & \raisebox{-.5\height}{\includegraphics[height=\exResultsHeight]{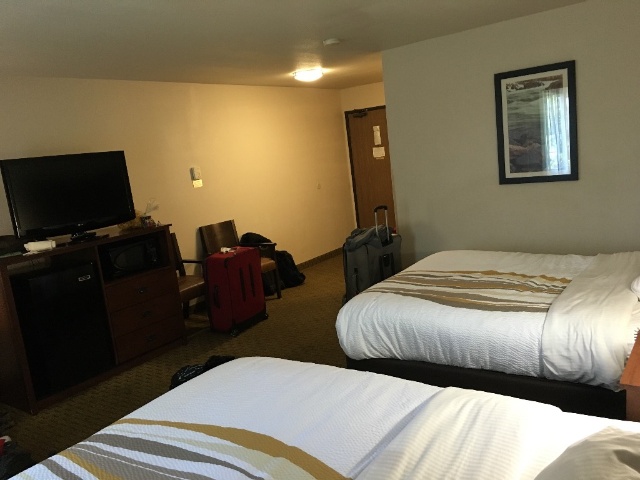}}
        &\raisebox{-.5\height}{\includegraphics[height=\exResultsHeight]{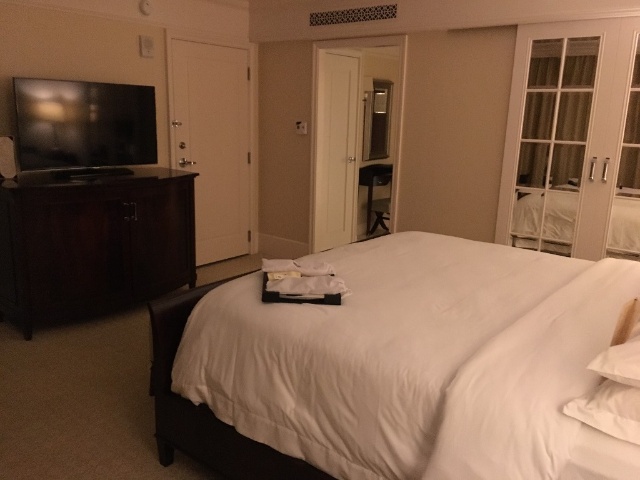}}
        &\raisebox{-.5\height}{\includegraphics[height=\exResultsHeight]{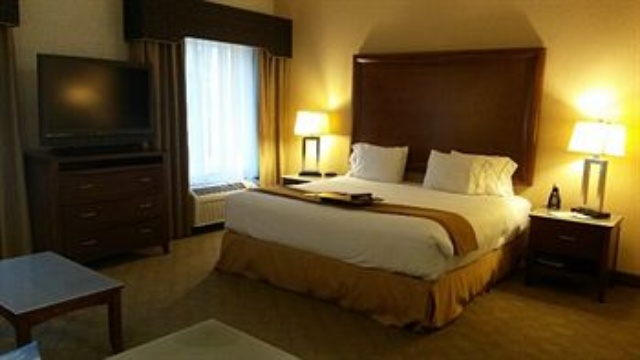}}
        &\raisebox{-.5\height}{\includegraphics[height=\exResultsHeight]{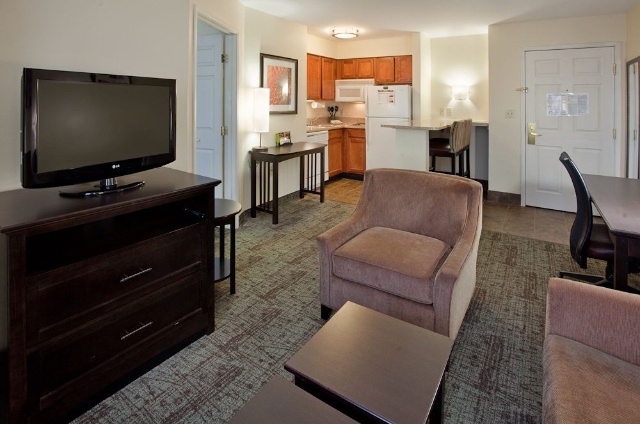}}
        &\raisebox{-.5\height}{\includegraphics[height=\exResultsHeight]{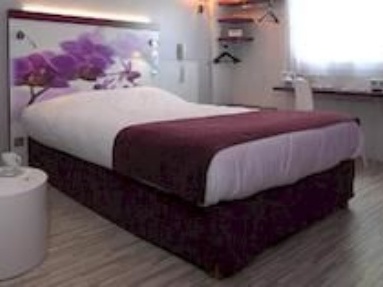}}
    \\
     &
        \raisebox{-.5\height}{{\sc Fixed-Scene}}
        &\raisebox{-.5\height}{\includegraphics[height=\exResultsHeight]{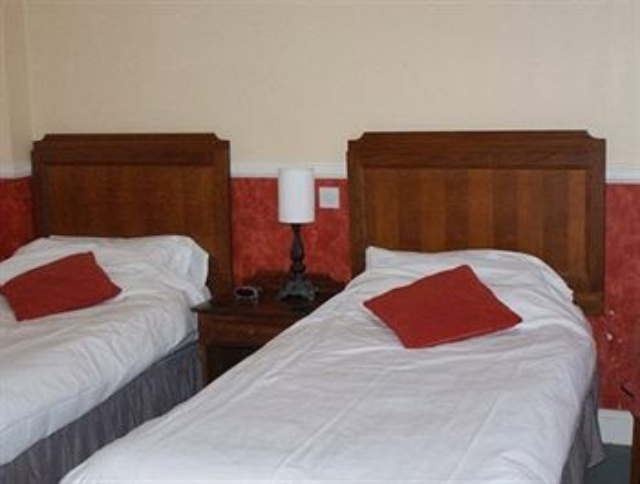}}
        &\raisebox{-.5\height}{\includegraphics[height=\exResultsHeight]{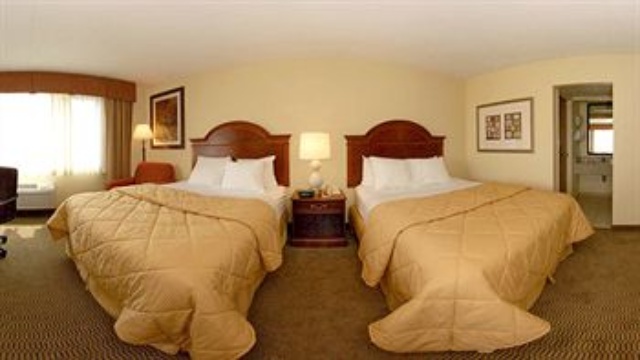}}
        &\raisebox{-.5\height}{\includegraphics[height=\exResultsHeight]{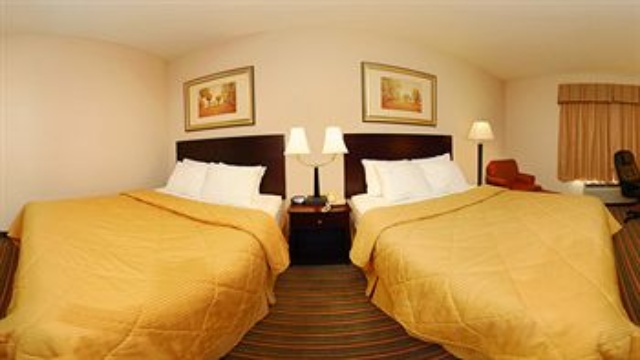}}
        &\raisebox{-.5\height}{\includegraphics[height=\exResultsHeight]{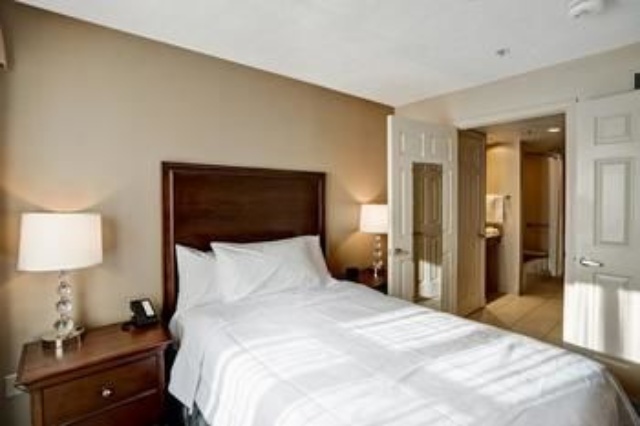}}
        &\raisebox{-.5\height}{\includegraphics[height=\exResultsHeight]{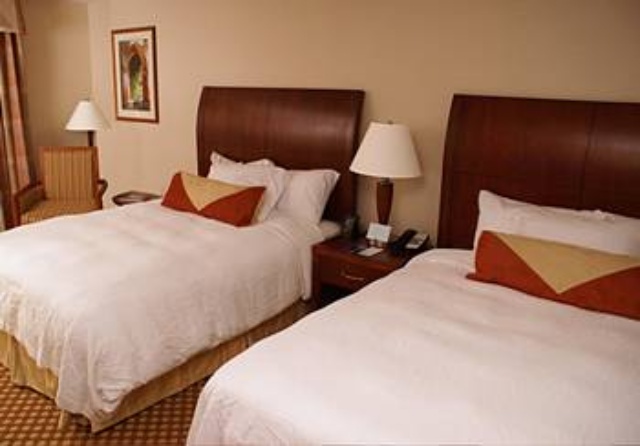}}
    \\
    &
        \raisebox{-.5\height}{Ours}
        & \raisebox{-.5\height}{\includegraphics[height=\exResultsHeight]{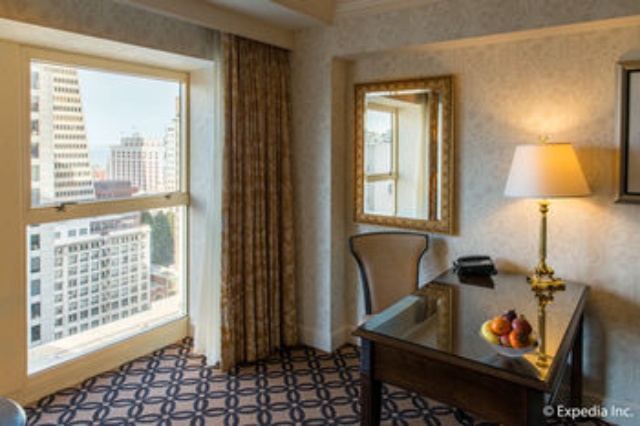}}
        &\raisebox{-.5\height}{\fcolorbox{green}{green}{\includegraphics[height=\exResultsHeight]{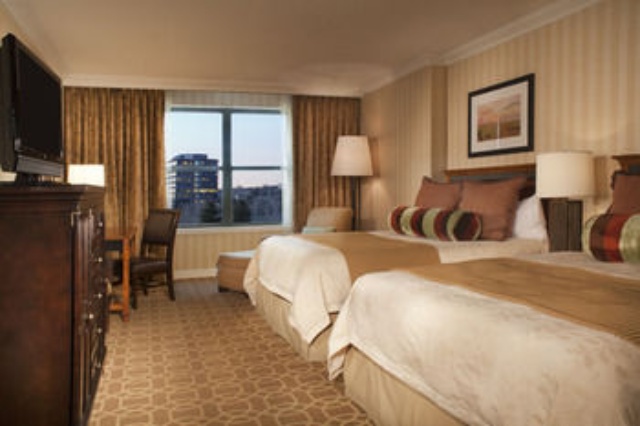}}}
        &\raisebox{-.5\height}{\fcolorbox{green}{green}{\includegraphics[height=\exResultsHeight]{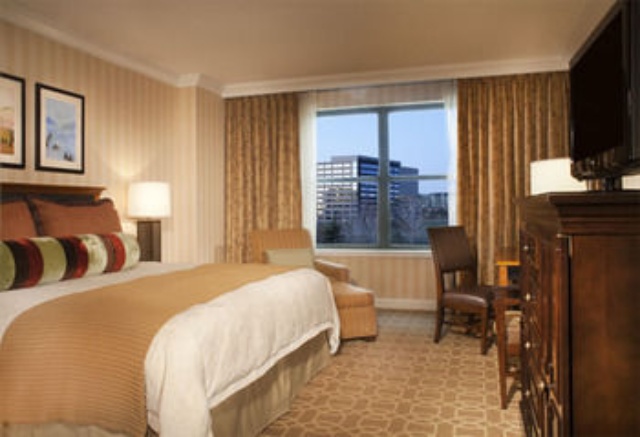}}}
        &\raisebox{-.5\height}{\fcolorbox{green}{green}{\includegraphics[height=\exResultsHeight]{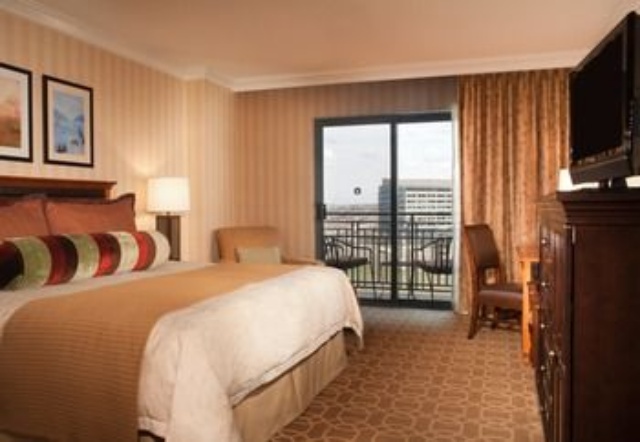}}}
        & \raisebox{-.5\height}{\includegraphics[height=\exResultsHeight]{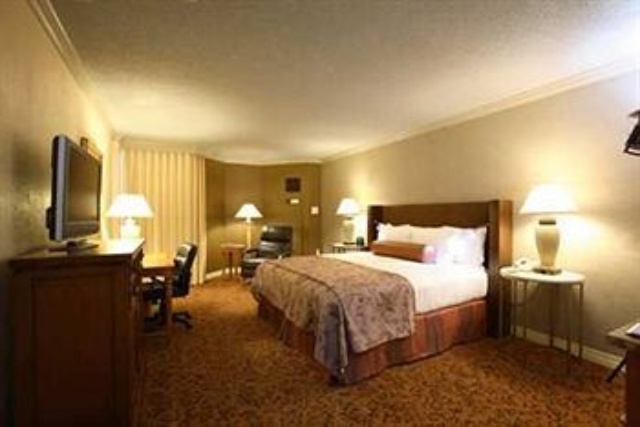}}\\
        
    \cline{1-7}  
    \multirow{3}{*}{\raisebox{-1.2\height}{\includegraphics[width=\queryImWidth]{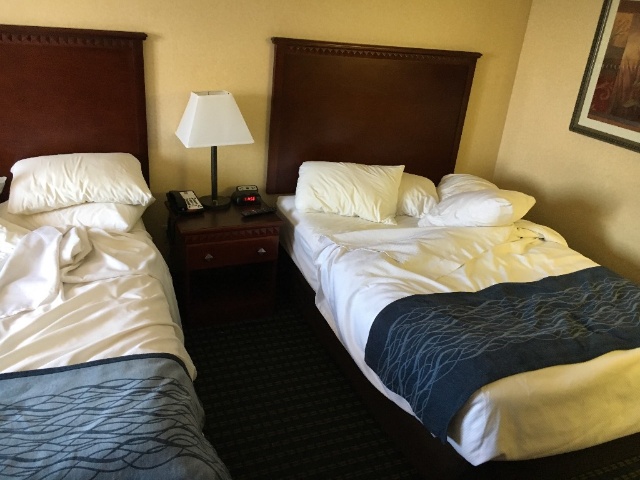}}}
        &\raisebox{-.5\height}{{\sc Fixed-Object}}
        & \raisebox{-.5\height}{\includegraphics[height=\exResultsHeight]{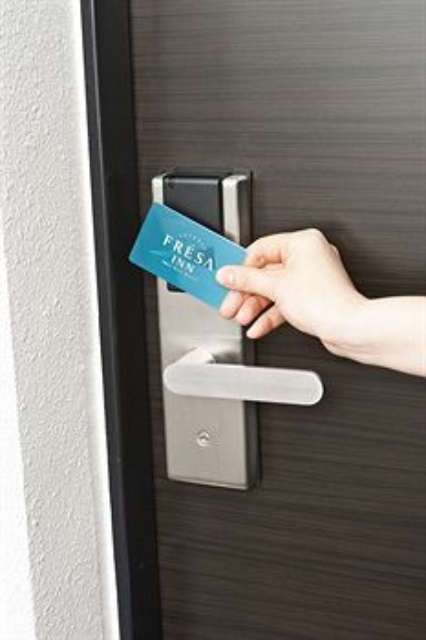}}
        &\raisebox{-.5\height}{\includegraphics[height=\exResultsHeight]{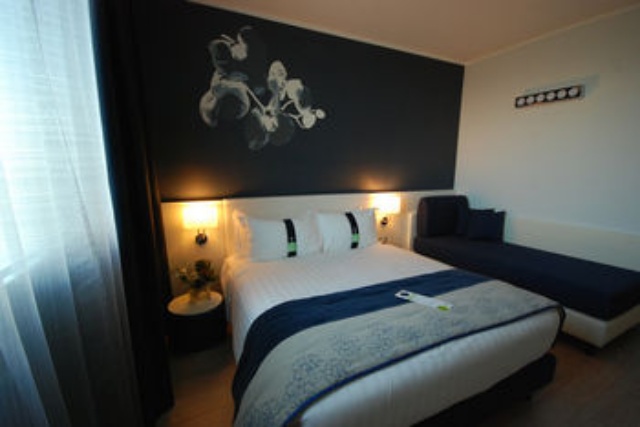}}
        &\raisebox{-.5\height}{\includegraphics[height=\exResultsHeight]{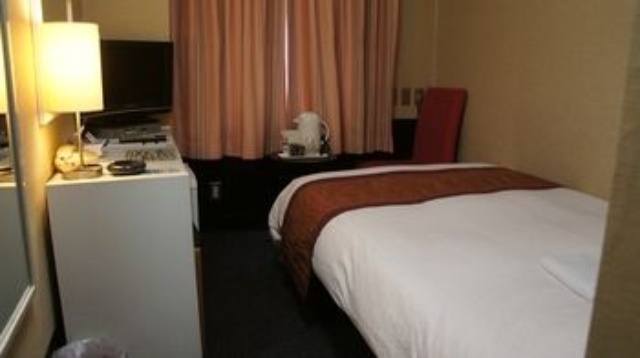}}
        &\raisebox{-.5\height}{\includegraphics[height=\exResultsHeight]{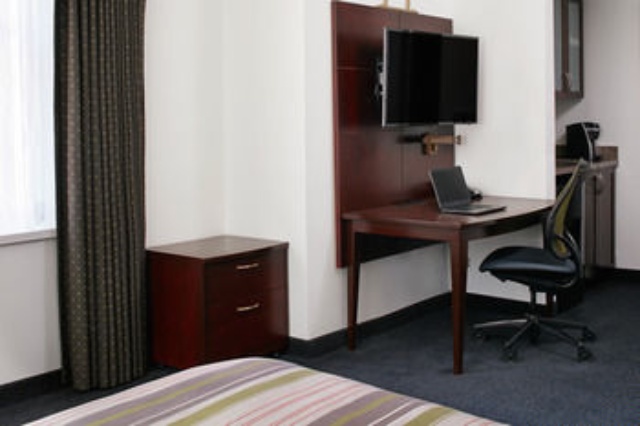}}
        &\raisebox{-.5\height}{\includegraphics[height=\exResultsHeight]{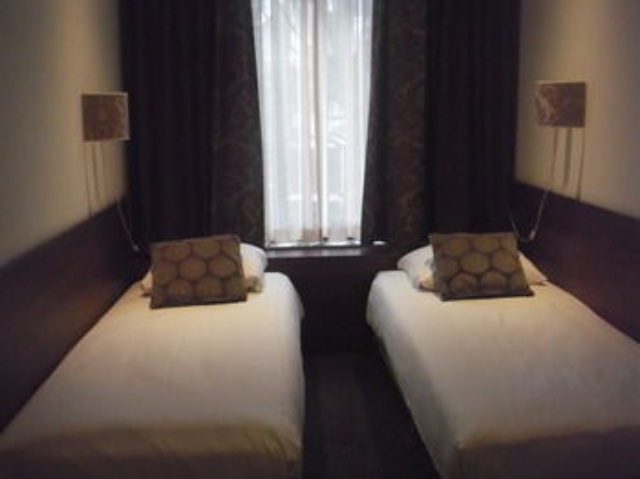}}
    \\
        &
        \raisebox{-.5\height}{{\sc Fixed-Scene}}
        &\raisebox{-.5\height}{\includegraphics[height=\exResultsHeight]{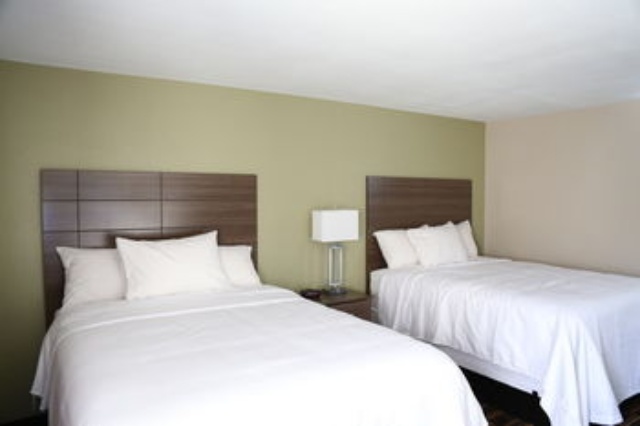}}
        &\raisebox{-.5\height}{\includegraphics[height=\exResultsHeight]{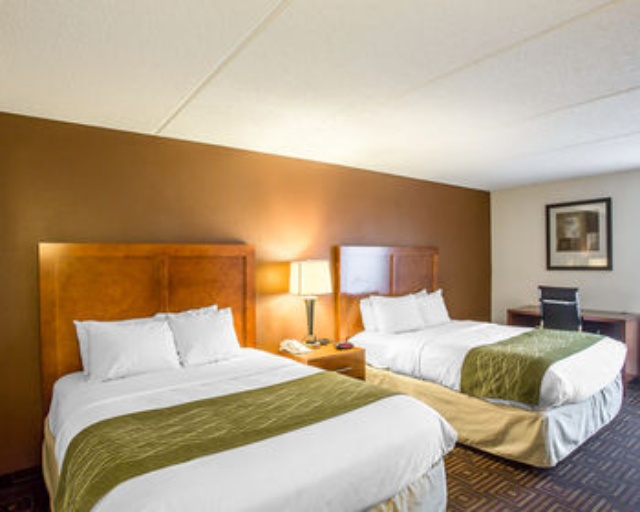}}
        &\raisebox{-.5\height}{\includegraphics[height=\exResultsHeight]{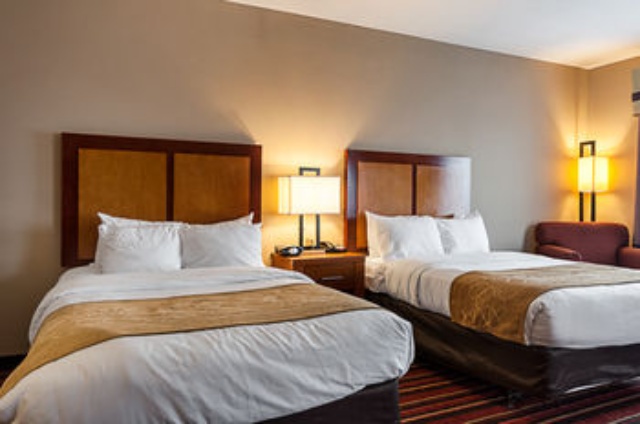}}
        &\raisebox{-.5\height}{\includegraphics[height=\exResultsHeight]{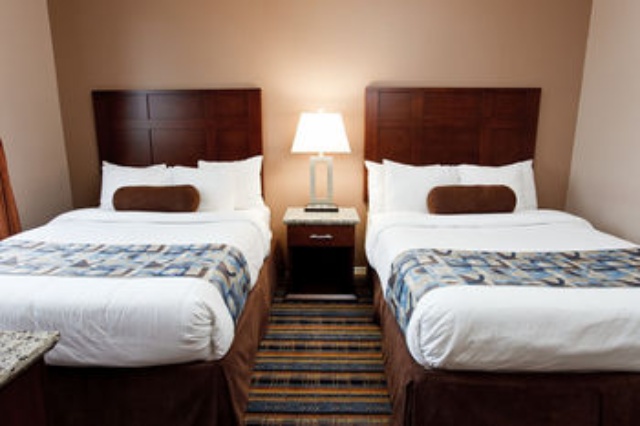}}
        &\raisebox{-.5\height}{\includegraphics[height=\exResultsHeight]{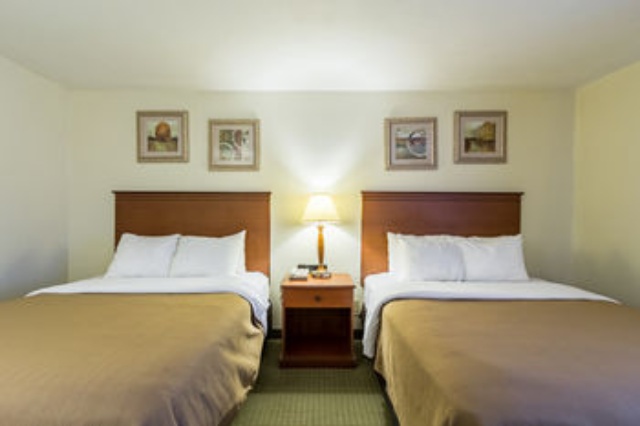}}
    \\
        &
        \raisebox{-.5\height}{Ours}
        & \raisebox{-.5\height}{\includegraphics[height=\exResultsHeight]{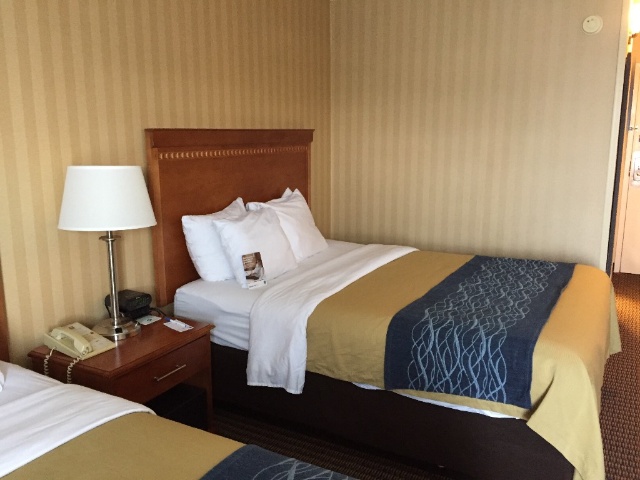}}
        &\raisebox{-.5\height}{\includegraphics[height=\exResultsHeight]{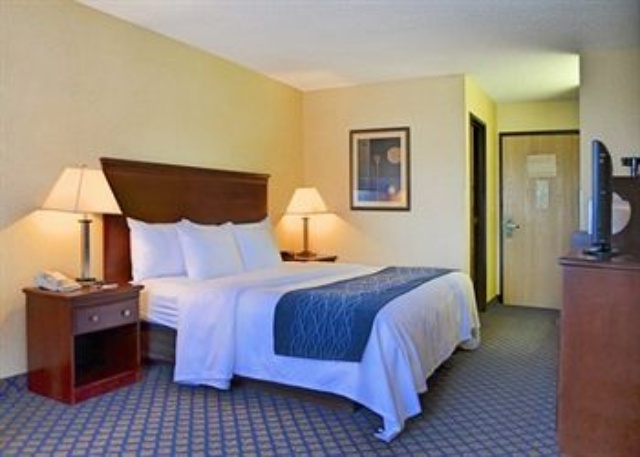}}
        &\raisebox{-.5\height}{\includegraphics[height=\exResultsHeight]{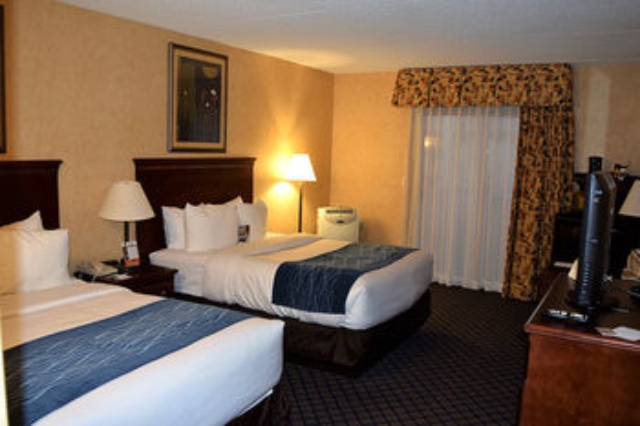}}
        &\raisebox{-.5\height}{\includegraphics[height=\exResultsHeight]{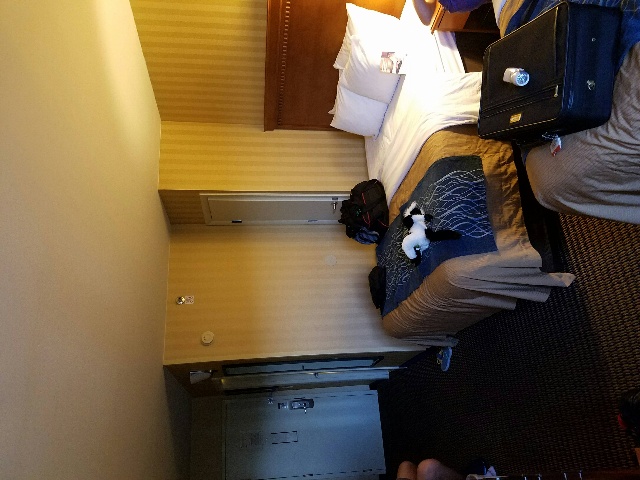}}
        & \raisebox{-.5\height}{\fcolorbox{green}{green}{\includegraphics[height=\exResultsHeight]{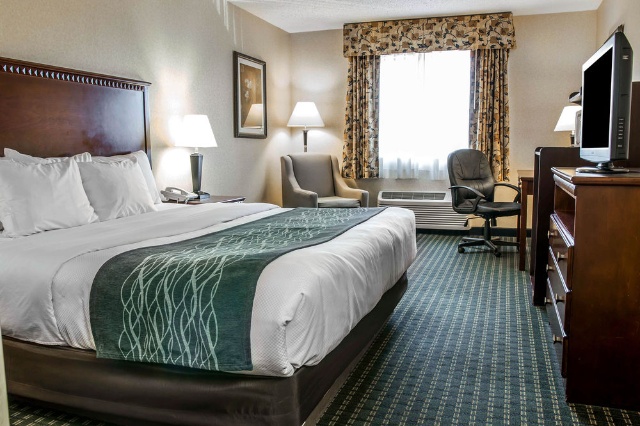}}}\\
    \end{tabular}
    \caption{The top 5 most similar results for the models trained on the Places-365 dataset, the ILSVRC dataset, and our model trained on travel website and TraffickCam images with data augmentation. Images from the correct hotel instance are highlighted in green.}
    \label{fig:top5_results}
\end{figure*}

Figure~\ref{fig:top5_results} shows the top 5 results for several query images using {\sc Fixed-Object}, {\sc Fixed-Scene} and our approaches.  Unlike {\sc Fixed-Object} and {\sc Fixed-Scene}, our model appears to encode information about the important colors and objects in a hotel room. In the top example in Figure~\ref{fig:top5_results}, our approach finds examples from the correct hotel, as well as other images with similar blue walls and headboards. Our model also performs reasonably well even in the case where there is large amounts of clutter in the query image, as seen in the middle example in Figure~\ref{fig:top5_results}. The last example in Figure~\ref{fig:top5_results} highlights the difficulty of hotel instance recognition given the similarity between instances of the same hotel chain -- nearly all of the top images retrieved by our model are from the correct hotel chain, but
not necessarily the correct hotel. 

\subsection{Classification}
For the classification task, we adapt the image embedding approaches used for image retrieval to report class posterior probabilities. For each method for each test image, we find the 1000 most similar images in the database using cosine similarity between the output features. The proportion of each class (hotel instance or hotel chain)
in the resulting set is the estimate of the posterior probability.

\begin{table}
    \def\arraystretch{1.1}
    \centering
    \begin{tabular}{c|c|c|c|c|}
        \multicolumn{1}{r}{\textbf{Occlusion:}} & \multicolumn{1}{c}{\textbf{none}} & \multicolumn{1}{c}{\textbf{low}} & \multicolumn{1}{c}{\textbf{medium}} & \multicolumn{1}{c}{\textbf{high}} \\
        \cline{1-5}
        \multicolumn{1}{|l|}{{\sc Fixed-Object}} & 34.1 & 34.3 & 34.5 & 34.4 \\
        \multicolumn{1}{|l|}{{\sc Fixed-Scene}} & 33.8 & 33.9 & 34.1 & 34.2 \\
        \multicolumn{1}{|l|}{Ours} & \textbf{23.8} & \textbf{24.0}  & \textbf{25.4} & \textbf{27.2} \\
        \cline{1-5}
    \end{tabular}
    \caption{Multi-class log loss for
    each method on the hotel instance classification task. }
    \label{tab:log_loss}
\end{table}

Table~\ref{tab:log_loss} shows the multiclass log loss for each method for varying levels of occlusions in the test images. 
In all cases, our approach outperforms features from the pretrained models. However, there is still significant room 
for improved classification performance. 

\subsection{Ablation Study}
To quantify the effects of both the inclusion of the crowdsourced data and the augmentation steps in our approach, we compare the results of variants of our method on the hotel instance retrieval task with and without significant occlusions.

\begin{table}
    \centering
    %   \begin{tabular}{c|ccc|}
    %     \multicolumn{1}{c}{} & \multicolumn{3}{c}{\textbf{no occlusions}} \\
    %     \cline{2-4}
    %     \multicolumn{1}{c|}{\textbf{Model}} & K=1 & 3 & 5 \\
    %     \cline{1-4}
    %     \multicolumn{1}{|c|}{Method$_1$} & 27.3 & 59.1 & 90.9 \\
    %     \multicolumn{1}{|c|}{Method$_2$} & 39.5 & 67.5 & 89.3\\
    %     \multicolumn{1}{|c|}{Method$_3$} & \textbf{39.7} & \textbf{68.6} & \textbf{90.6}\\
    %     \cline{1-4}
    %     \multicolumn{4}{c}{}\\
    %     \multicolumn{4}{c}{(b) By hotel chain.}
    % \end{tabular}
    % \quad
    \begin{tabular}{c|ccc|ccc|}
        \multicolumn{1}{r}{\textbf{Occlusion:}} & \multicolumn{3}{c}{\textbf{none}} & \multicolumn{3}{c}{\textbf{medium}} \\
        \cline{2-7}
        \multicolumn{1}{c|}{} & K=1 & 10 & 100 & 1 & 10 & 100 \\
        \cline{1-7}
        \multicolumn{1}{|l|}{Ours -A,-I} & 4.7 & 9.6 & 20.0 & 1.8 & 4.0 & 9.4\\
        \multicolumn{1}{|l|}{Ours -A} & \textbf{8.1} & \textbf{18.4} & \textbf{36.0} & 3.5 & 9.2 & 12.8 \\
        \multicolumn{1}{|l|}{Ours} & \textbf{8.1} & 17.6 & 34.8 & \textbf{5.9} & \textbf{14.1} & \textbf{29.9}\\
        \cline{1-7}
    \end{tabular}
    
    \caption{Ablation study reported as top-$K$ hotel instance retrieval for our method and variants without data augmentation (-A) and without crowdsourced images (-I).}
    \label{tab:ablationStudy}
\end{table}

Table~\ref{tab:ablationStudy} shows the results for the ablation experiment. We evaluate our approach without the data augmentation steps and additionally without including the crowdsourced images, which are those most similar to the real-world images. The inclusion of the crowdsourced images has a significant impact on the performance both with and without occlusions in the test image. The data augmentation steps do not have an impact on the performance in the un-occluded cases, but in the medium occlusion case, which roughly corresponds to sizes of the masked regions in real-world cases, the benefits of the data augmentation steps are apparent, increasing the top-$K$ accuracy by more than 50\% for $K=10$.

\section{Conclusion}
In this paper, we introduced Hotels-50K, a dataset of over a million images of hotel rooms from 50,000 different hotels around the world. This dataset should further the state of the art in hotel recognition from images. We present an approach trained on the Hotels-50K dataset that outperforms fixed features from generic object and scene models. The Hotels-50K dataset, pre-trained models and code to replicate our baseline approaches can be found at \url{https://github.com/GWUvision/Hotels-50K}. The baseline approach is currently deployed for use by human trafficking investigators, including the National Center for Missing and Exploited Children, and novel algorithms can be quickly deployed to improve search performance in ongoing investigations.

This project is based in part on work supported through the National Institute of Justice (Grant 2018-75-CX-0038) and a gift from Adobe Inc.

%References and End of Paper
%These lines must be placed at the end of your paper
\bibliography{bib}
\bibliographystyle{aaai}
\end{document}